\documentclass[10pt,twocolumn,letterpaper]{article}

\usepackage{cvpr}              

\definecolor{r1}{RGB}{220, 100, 100}
\definecolor{r2}{RGB}{80, 80, 240}
\definecolor{r3}{RGB}{0, 180, 0}

\newcommand{\blue}[1]{{\color{blue}#1}}
\newcommand{\gray}[1]{{\color{black!50}#1}}

\usepackage{graphicx}
\usepackage{microtype}
\usepackage{amssymb}
\usepackage[skip=2pt]{caption}

\usepackage{array} 
\usepackage{multirow}
\usepackage{stfloats} 
\usepackage{caption}

\usepackage{listings}
\usepackage{color}
\usepackage[most]{tcolorbox}

\definecolor{cvprblue}{rgb}{0.21,0.49,0.74}
\usepackage[pagebackref,breaklinks,colorlinks,allcolors=cvprblue]{hyperref}

\def\confName{CVPR}
\def\confYear{2026}

\title{OV-Stitcher: A Global Context-Aware Framework \\ for Training-Free Open-Vocabulary Semantic Segmentation}

\author{
Seungjae Moon\qquad Seunghyun Oh\qquad Youngmin Ro\textsuperscript{*} \\
Machine Intelligence Laboratory, University of Seoul, Korea \\
\tt\small \{msj0243, osh1795, youngmin.ro\}@uos.ac.kr \\
\tt\small \url{https://github.com/atw617/OV-Stitcher}
}

\newcommand{\second}[1]{{\underline{#1}}}

\usepackage{lipsum}
\usepackage{array}
\usepackage{times}
\usepackage{epsfig}
\usepackage{graphicx}
\usepackage{float}
\usepackage{wrapfig}
\usepackage{amsmath,amssymb,amsthm}
\usepackage{algorithm,algorithmicx,algpseudocode}
\usepackage{bm,xspace}
\usepackage{comment}
\usepackage{multirow}
\usepackage{balance}
\usepackage{url}
\usepackage{booktabs}
\usepackage{etoolbox,siunitx}
\usepackage{calc}
\usepackage{pifont,hologo}
\usepackage{color}
\usepackage{glossaries}
\usepackage[table]{xcolor}
\usepackage{adjustbox}
\usepackage{amsmath}
\usepackage{enumitem}
\usepackage{bbding}  %
\usepackage[normalem]{ulem}  %
\usepackage{contour}
\usepackage{soul}%
\usepackage{tocloft} %
\usepackage{etoc} %
\definecolor{cvprblue}{rgb}{0.21,0.49,0.74}
\usepackage[pagebackref,breaklinks,colorlinks,allcolors=cvprblue]{hyperref}
\usepackage{natbib}
\PassOptionsToPackage{square,numbers,comma,sort}{natbib}

\newacronym{ovss}{OVSS}{Open vocalbualry semantic segmentation}
\newacronym{fsad}{FSAD}{Few-Shot Anomaly Detection}
\newacronym{zsad}{ZSAD}{Zero-Shot Anomaly Detection}
\newacronym{mcad}{MCAD}{Multi-Class Anomaly Detection}

\newcommand{\reffig}[1]{Fig.~\ref{#1}}

\definecolor{blue}{HTML}{004bb3}

\def\@onedot{\ifx\@let@token.\else.\null\fi\xspace}

\DeclareMathAlphabet{\mathsfit}{\encodingdefault}{\sfdefault}{m}{sl}
\SetMathAlphabet{\mathsfit}{bold}{\encodingdefault}{\sfdefault}{bx}{n}

\let\originalleft\left
\let\originalright\right
\renewcommand{\left}{\mathopen{}\mathclose\bgroup\originalleft}
\renewcommand{\right}{\aftergroup\egroup\originalright}

\definecolor{codegreen}{rgb}{0,0.6,0} 

\lstdefinestyle{mystyle}{
    backgroundcolor=\color{white}, 
    commentstyle=\color{codegreen}, 
    keywordstyle=\color{blue}, 
    stringstyle=\color{codegreen}, 
    basicstyle=\ttfamily\footnotesize, 
    breaklines=true, 
    captionpos=b, 
    numbers=left, 
    keepspaces=true }
\begin{document}

\maketitle
\renewcommand{\thefootnote}{\fnsymbol{footnote}}
\footnotetext[1]{Corresponding author.}
\renewcommand{\thefootnote}{\arabic{footnote}}

\begin{abstract}
Training-free open-vocabulary semantic segmentation~(TF-OVSS) has recently attracted attention for its ability to perform dense prediction by leveraging the pretrained knowledge of large vision and vision–language models, without requiring additional training.
However, due to the limited input resolution of these pretrained encoders, existing TF-OVSS methods commonly adopt a sliding-window strategy that processes cropped sub-images independently.
While effective for managing high-resolution inputs, this approach prevents global attention over the full image, leading to fragmented feature representations and limited contextual reasoning.
We propose OV-Stitcher, a training-free framework that addresses this limitation by stitching fragmented sub-image features directly within the final encoder block.
By reconstructing attention representations from fragmented sub-image features, OV-Stitcher enables global attention within the final encoder block, producing coherent context aggregation and spatially consistent, semantically aligned segmentation maps.
Extensive evaluations across eight benchmarks demonstrate that OV-Stitcher establishes a scalable and effective solution for open-vocabulary segmentation, achieving a notable improvement in mean Intersection over Union~(mIoU) from 48.7 to 50.7 compared with prior training-free baselines.

\end{abstract}

\vspace{-10px}
\section{Introduction}
\label{sec:intro}

\begin{figure}[t]
  \centering
   \includegraphics[width=0.98\columnwidth]{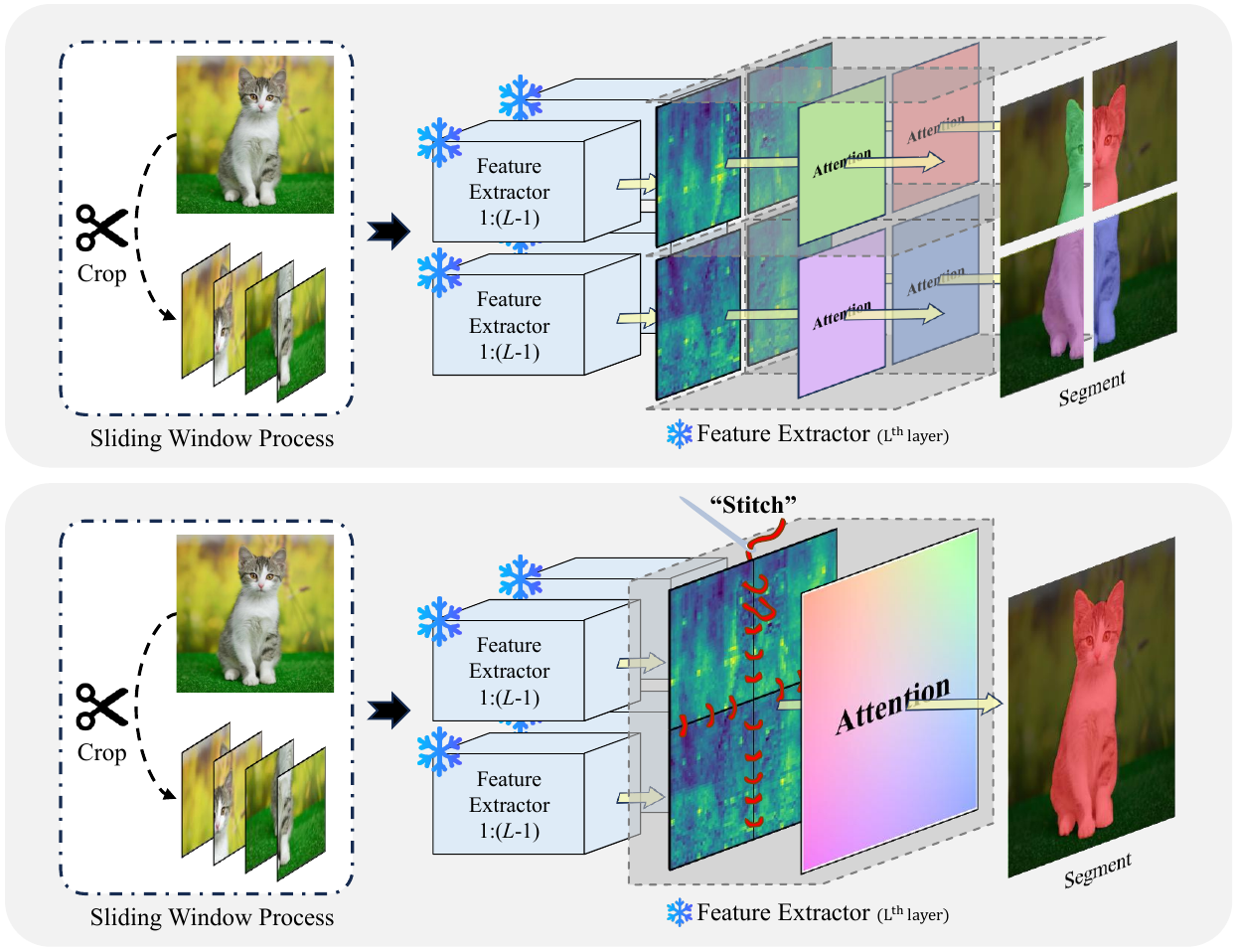}

   \caption{\textbf{Top}: Prior works process cropped sub-images independently, preventing attention across different sub-image features.
    \textbf{Bottom}: We introduce a Stitch Attention mechanism that enables global attention across all cropped regions, yielding more coherent and contextually consistent feature integration.}
    \vspace{-16px}
   \label{fig:teaser}
\end{figure}

Open-vocabulary semantic segmentation~(OVSS) seeks to assign pixel-level semantic labels guided by arbitrary text descriptions, rather than being limited to a fixed set of predefined categories. By leveraging the strong generalization ability of large-scale vision–language models (VLMs) such as CLIP~\cite{clip}, OVSS enables recognition and segmentation of novel concepts, reducing dependence on costly pixel-level human annotations while still benefiting from knowledge learned during large-scale pretraining, thereby allowing flexible adaptation across diverse domains. Within this paradigm, training-free OVSS~(TF-OVSS) represents a particularly attractive direction: instead of requiring additional fine-tuning or task-specific supervision, TF-OVSS directly exploits the pretrained knowledge and strong generalization capacity of VLMs to perform dense prediction. This allows segmentation to be achieved purely from pretrained representations, demonstrating the full potential of vision–language alignment without the need for further training.

\begin{figure*}[t!]
    \centering
    \includegraphics[width=1\linewidth]{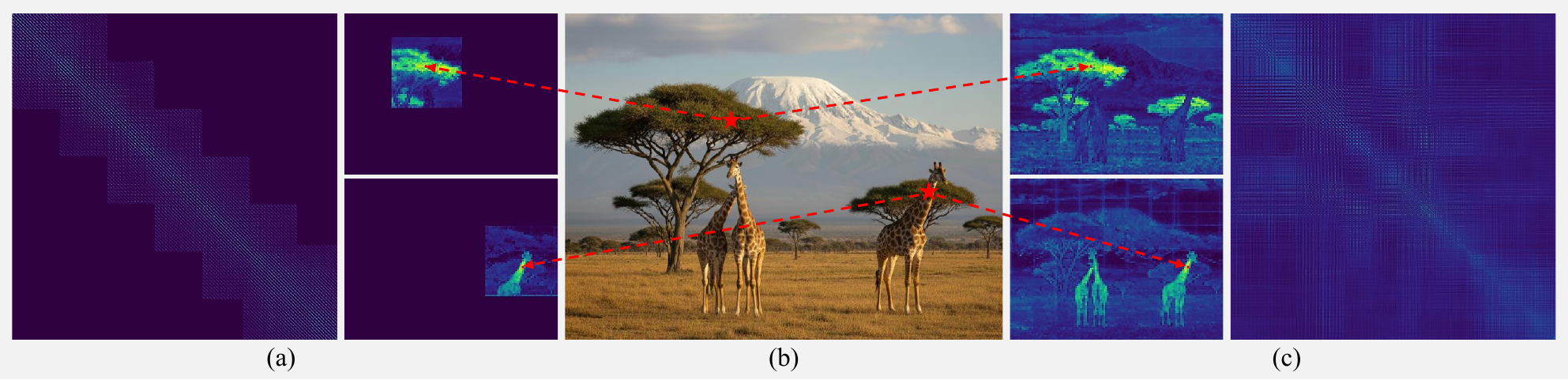}
    \caption{Illustration of the attention maps and patch-interactions for prior methods and our Stitch Attention. (a) presents prior methods, and (c) presents our approach.}
    \label{fig:analysis}
    \vspace{-0.5cm}
\end{figure*}

However, CLIP, as a vision–language model, is trained with an image-level contrastive objective, which encourages strong alignment between image-level representations and corresponding text descriptions.
While this enables effective recognition of diverse concepts, it does not explicitly provide pixel-level supervision, which poses challenges for directly applying CLIP for dense prediction tasks such as TF-OVSS. To address this issue, several training-free methods ~\cite{naclip, sclip, gem, sfp, clip_surgery, maskclip_tf} have been proposed to extract spatially variant features that better capture local semantics by modifying CLIP’s self-attention mechanism to yield more localized feature interactions.

Moreover, ProxyCLIP~\cite{proxyclip} utilizes spatial affinity information derived from vision foundation models (VFMs)~\cite{dino, dinov2, mae, sam}, which provides localization cues to enhance the correspondence between visual patches and text embeddings without directly modifying the text alignment.
These approaches generally focus on producing high-quality attention maps that can more accurately localize image regions, thereby strengthening patch-level semantics and improving overall segmentation performance.
More recently, several training-free approaches ~\cite{corrclip, textregion} have actively leveraged the Segment Anything Model (SAM) ~\cite{sam, sam2} to enhance open-vocabulary segmentation.
The methods utilize SAM’s mask-generation capability either to refine attention maps or for post-processing, producing more coherent and semantically consistent patch representations.
These approaches enhance the spatial precision of predictions and overall segmentation quality, highlighting the advantage of complementing vision–language models with both segmentation-oriented and representation-level vision foundation models.

\vspace{-1pt}

Despite these advances, current TF-OVSS approaches remain fundamentally constrained by the limited input resolution of CLIP.
To handle higher resolution inputs, existing methods typically adopt a \textit{sliding-window} strategy, where the source image is divided into multiple overlapping sub-images that are processed independently, and the resulting logits are subsequently stitched together to form the final prediction map.
While this approach effectively mitigates the resolution constraint, processing each sub-image independently limits interactions between them, which can lead to fragmented feature representations and the loss of global contextual information.
This phenomenon is reflected in the attention maps produced by existing sliding-window methods, where the lack of interaction between sub-images becomes apparent.
As shown in \reffig{fig:analysis} (a), each sub-image attends only to its own patches, without interacting with patches from other sub-images.
The resulting attention maps are fragmented, with regions remaining confined within each sub-image, failing to capture long-range dependencies and global context.
This fragmentation reveals the restricted receptive field inherent to sub-image processing and provides clear empirical evidence of the challenges associated with the sliding-window paradigm.
Consequently, the model’s ability to reason about relationships between distant regions is limited, which can lead to inconsistent segmentation predictions and reduced coherence across the image (see \S.~\ref{subsec:Analysis} for a detailed discussion).
This limitation is especially noticeable in scenes that are large-scale or complex, where the lack of global attention can hinder accurate alignment between visual patches and their corresponding semantic labels.
The inability to aggregate information across the entire image underscores the need for approaches capable of reconstructing global feature interactions while preserving the fine-grained information within each sub-image.
Motivated by these observations, we propose OV-Stitcher, a training-free, global context-aware framework that reconstructs feature interactions across sub-images.

At the core of OV-Stitcher lies Stitch Attention, a mechanism designed to overcome the fragmentation resulting from sliding-window processing.
In Fig.~\ref{fig:teaser}, we briefly show how conventional sliding-window methods and OV-Stitcher process the image, allowing their differences to be clearly observed.
Stitch Attention operates within the encoder block, stitching features across sub-images immediately before the attention computations. This design enables information exchange beyond local patch boundaries, bridging fragmented regions into unified representations. As a result, this design enables the model to capture long-range dependencies and global context, leading to more coherent and semantically consistent feature representations. In addition to Stitch Attention, to mitigate class ambiguities in large and coherent regions, OV-Stitcher incorporates class-biased text prompts. These prompts ensure a more reliable mapping between predicted segments and their corresponding text embeddings, reinforcing semantic alignment across sub-images. The synergistic design of these components enables OV-Stitcher to achieve state-of-the-art average performance across eight benchmarks.

Our contributions can be summarized as: 
\begin{itemize}
    \item We identify the challenges of applying sliding-window TF-OVSS approaches, highlighting the lack of attention between sub-images and the loss of global context caused by the sliding-window based processing.
    \item Motivated by this analysis, we propose OV-Stitcher, a training-free framework that reconstructs global feature interactions via Stitch Attention and incorporates class-biased text prompts to enhance semantic alignment.
    \item OV-Stitcher achieves state-of-the-art average performance across eight benchmarks, demonstrating the effectiveness of its synergistic design in improving open-vocabulary segmentation.
\end{itemize}
\section{Related Works}
\label{sec:2_relatedwork}
\subsection{Vision Language and Foundation Models}
\textbf{Vision-Language Models~(VLMs)}~\cite{clip,laionclip,dfnclip,metaclip,siglip} are multimodal architectures trained to align visual and textual representations in a unified embedding space. CLIP~\cite{clip}, a representative VLM, learns the rich correspondence between images and text through contrastive pre-training. This enables remarkable generalization performance on various downstream tasks, such as zero-shot classification, providing a crucial foundation for open-vocabulary capabilities. 

\noindent\textbf{Vision Foundation Models~(VFMs)}~\cite{dino,dinov2,dinov3,vit_reg,mae,sim_mim,i_jepa} learn general  and transferable visual representations from large-scale, diverse data. Their representations jointly capture semantic information and spatial details, remaining robust across scales and contexts. Therefore, VFMs deliver consistent gains across a broad range of downstream tasks, including classification~\cite{dino_classification}, detection~\cite{vfm_od} and segmentation~\cite{soma, rein}. Self-supervised VFMs learn the intrinsic structure and patterns of images without labels, yielding highly generalizable feature spaces. Notably, DINO~\cite{dino,dinov2,dinov3,vit_reg} produces semantically organized embeddings and exhibits strong cross-domain generalization. In addition, the Segment Anything Model~(SAM)~\cite{sam,sam2} enables prompt-based zero-shot segmentation and produces precise masks irrespective of class. SAM’s rich spatial representations are effectively leveraged in a variety of dense prediction scenarios~\cite{change_det, sam_r1}.

\subsection{Open-Vocabulary Semantic Segmentation}
Open-Vocabulary Semantic Segmentation (OVSS) aims to perform pixel-level semantic segmentation for arbitrary concepts described by natural language, beyond a predefined set of categories.
Training-based methods typically build upon CLIP and fine-tune the model using additional mask ~\cite{train1_liang2023open,cat_seg,train2_liu2023open,train3_use,train4_sed,train5_san,train6_fcclip}, textual descriptions~\cite{text1,text2_SegCLIP,text3_image,text4_rewrite,text5_uncovering}, or knowledge distillation procedures~\cite{clip_dinoiser} to achieve dataset-specific optimization.
However, these approaches depend on large-scale labeled data and can partially compromise the inherent open-vocabulary generalization capability of CLIP~\cite{car, corrclip}.

In contrast, training-free approaches~\cite{sclip, cliptrase, gem, lavg, clearclip, clip_surgery, corrclip, sfp, maskclip_tf} enable dense prediction without additional training by modifying CLIP's architecture or integrating external representations. These approaches primarily focus on mitigating the localization limitations of CLIP, namely the lack of patch-level spatial alignment resulting from its image-level supervision~\cite{clip_dinoiser,memba_as_bridge}. For instance, studies have proposed enhancing local semantic consistency by transforming CLIP's query-key attention into forms of self-self attention (e.g., value-value, key-key)~\cite{sclip,sc_clip,gem,clearclip,clip_surgery}.

Furthermore, ProxyCLIP~\cite{proxyclip} enhances both semantic coherence and spatial consistency by combining CLIP with VFM representations, using DINO to strengthen local patch-level alignment.
Building on this foundation, several studies~\cite{corrclip,textregion} have incorporated SAM~\cite{sam,sam2}, leveraging its mask-generation capability to provide spatial cues and post-processing, achieving more precise localization and coherent segmentation boundaries.

These methods typically handle high-resolution inputs by segmenting each sub-image individually using a sliding-window strategy. Our method further enables interactions across sub-images within the encoder layers, producing globally context-aware features that yield more coherent and consistent segmentations.
\section{Method}

\subsection{Preliminaries}
\label{subsec:preliminaries}
\noindent
\textbf{Similarity-Based Segmentation with Sliding-Window Inference.} 
In TF-OVSS, the limited input resolution of frozen backbones requires processing the image in a sliding-window manner. 
An input image $I$ is divided into $C$ overlapping crops $\{\tilde{I}_i \}_{i=1}^{C}$, and each crop is independently encoded by the vision encoder to obtain a local image feature map $\tilde{F}_{img}^{i, L}$ from last layer \textit{L}.
For each window, the segmentation logits are computed by measuring the similarity between the projected image features and the text embeddings of target categories:
\begin{equation}
    \tilde{Z}_i = \text{Proj}(\tilde{F}_{\text{img}}^{i, L}) F_{\text{text}}^\mathrm{\top},
\end{equation}
where $\text{Proj}(\cdot)$ aligns the visual features with the text feature space.
The local logits $\{\tilde{Z}_i \}^{C}_{i=1}$ are then spatially stitched through a stitching function $\mathcal{G}(\cdot)$ (implicitly followed by upsampling to the full image resolution) to reconstruct a full-resolution logit map:
\begin{equation}
    Z = \mathcal{G}(\{ \tilde{Z}_i \}_{i=1}^{C}).
\end{equation}
Finally, the semantic segmentation prediction is obtained by taking the class-wise maximum over the aggregated logits:
\begin{equation}
    pred = \mathop{\arg\max}\limits_{c}(Z).
\end{equation}
This formulation enables dense, training-free segmentation by integrating similarity-based predictions from local sliding windows into a high-resolution prediction.

However, since logits are computed independently for each crop, the sliding-window approach limits global interactions. 
Stitch Attention addresses this by integrating information across all crops at the last layer, producing more coherent features.

\noindent\textbf{Attention Map via Feature Affinity.} Recent approaches \cite{proxyclip, trident, corrclip} leveraging Vision Foundation Models (VFMs) have shown that high-quality visual features can effectively guide CLIP-based attention mechanisms.
Following this line of work, the attention map can be formulated based on feature similarity.
Given a feature map \(F_{\text{img}}\), a normalized self-similarity matrix, referred to as the affinity map, is computed as:
\begin{equation}
S = \frac{F_{\text{img}}}{\|F_\text{img}\|} \left(\frac{F_\text{img}} {\|F_\text{img}\|}\right)^\top,
\end{equation}
which captures the pairwise similarity between spatial features.
The attention map \(A\) is then defined as:
\begin{equation}
A = \text{Softmax}(S + M),
\end{equation}
where \(M\) is a mask highlighting relevant feature correlations.
This attention map \(A\) can then be applied to other feature representations, such as the value features from a CLIP encoder, via matrix multiplication. 
When applying spatially rich, high-quality features to construct the affinity map, this attention formulation can enhance the correspondence between spatial regions and downstream embeddings.

In our method, the affinity map is constructed from the features produced by our proposed Stitch Attention mechanism, with the mask \(M\) provided by SAM2~\cite{sam2}, following the implementation approach proposed in CorrCLIP~\cite{corrclip}.

\subsection{Analysis for Existing Approach}
\label{subsec:Analysis}
Recent training-free open-vocabulary segmentation methods have significantly enhanced the local perception capability of CLIP-based models, often employing a sliding-window strategy to handle higher-resolution images. While this approach effectively increases local recognition accuracy, it introduces an inherent limitation: each sub-image is encoded independently, so attention is applied only among tokens within the same sub-image, ignoring relationships across sub-images. Fig.~\ref{fig:analysis} (a) shows this limitation, highlighting how tokens from different sub-images do not interact. As a result, the global semantic coherence of objects throughout the image can be disrupted.

\begin{figure} 
    \centering
    \includegraphics[width=0.45\textwidth]{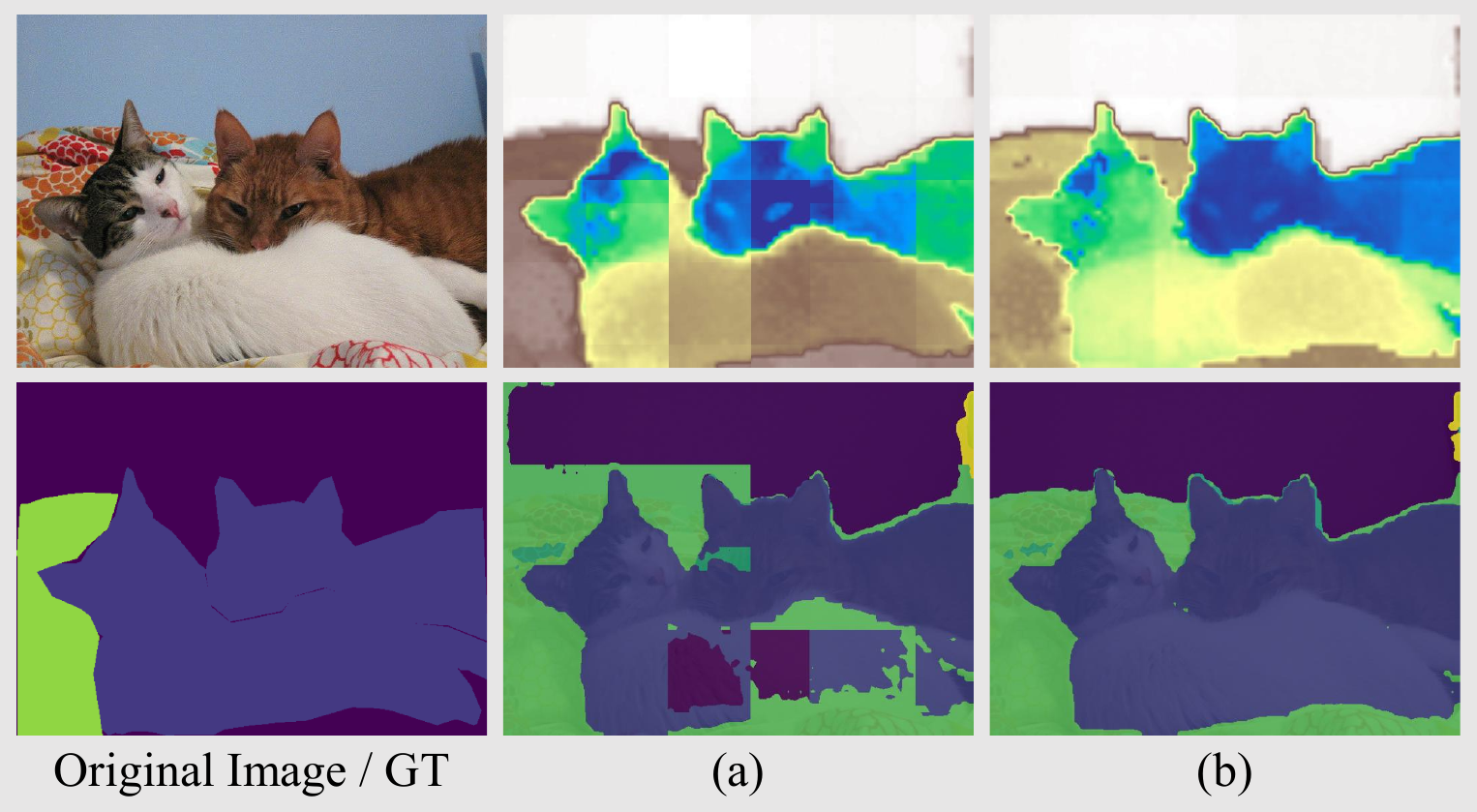}
    \caption{\textbf{Visualization of feature representations and segmentation results.} (a) and (b) show the image feature maps and segmentation results obtained from the baseline and the proposed Stitch Attention, respectively. The top row shows the feature maps after applying PCA, and the bottom row presents the corresponding segmentation results.}
    \label{fig:PCA}
    \vspace{-0.1cm}
\end{figure}
To investigate this issue, we visualize the feature map obtained from each independently encoded sub-image.
After reconstructing the global feature map by stitching these sub-image features, we apply PCA for qualitative analysis.
As shown in Fig.~\ref{fig:PCA} (top, a), the visualization reveals a fragmented feature structure, where even regions belonging to the same object show inconsistent representations, indicating that the encoding varies across sub-image boundaries.
The predicted segmentation result in Fig.~\ref{fig:PCA} (bottom, a) reflects the same inconsistency observed in the feature map.
These observations indicate that the independent encoding of sub-images leads to sub-optimal predictions, corroborating the limitations highlighted by the feature visualization in Fig.~\ref{fig:PCA} and the attention analysis in Fig.~\ref{fig:analysis}.
By comparison, Fig.~\ref{fig:PCA} (b) shows that the method proposed in \S.~\ref{subsec:stitch_attention} produces feature maps that are more structured and coherent across sub-image boundaries, which in turn leads to more consistent and accurate segmentation predictions.

\begin{figure*}
  \centering
  \includegraphics[width=1\textwidth]{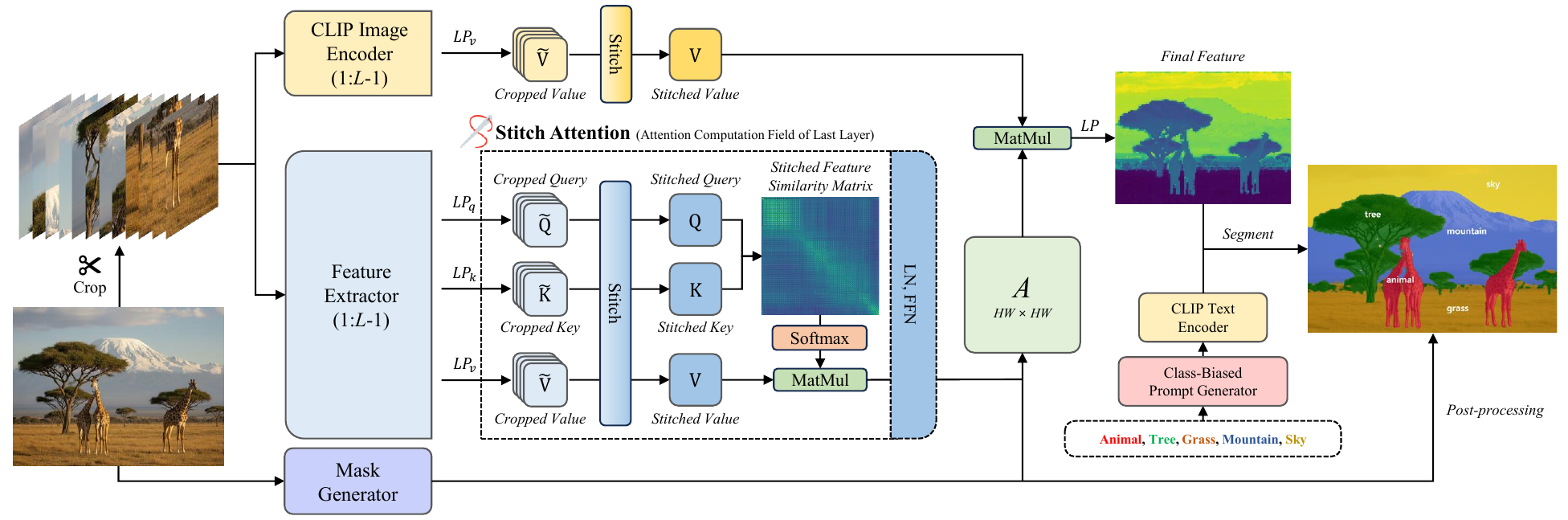}
  \caption{Overview of our method \textbf{OV-Stitcher}. Our core framework starts from processing each sub-image using a sliding window approach. From the final layer of each sub-image, we extract $\tilde{Q}$, $\tilde{K}$, and $\tilde{V}$ features, and stitch each type separately across all sub-images to form the global $Q$, $K$, and $V$. Self-attention on these stitched features produces a feature map capturing global correlations. The features resulting from Stitch Attention provide CLIP with global spatial information, enabling coherent reasoning across the full image. Additionally, we add \textbf{Class-Biased Prompts} to the existing prompts to generate text embeddings reducing ambiguity among similar categories.
}
  \label{overview}
  \vspace{-2mm}
\end{figure*}

\subsection{Stitch Attention}
\label{subsec:stitch_attention}
After analyzing the limitations of prior approaches (\S.~\ref{subsec:Analysis}), we introduce our Stitch Attention mechanism, which explicitly enhances the consistency of visual features across sub-images. Conventional sliding-window inference confines self-attention to each crop, hindering the modeling of dependencies across crop boundaries. Our method overcomes this limitation by stitching crop-level features into a single global representation before applying attention.

At the last encoder layer, the model generates query~$Q$, key~$K$, and value~$V$ $\in \mathbb{R^{\textit{C}\times \textit{hw}\times \textit{d}}}$ embeddings via linear projections, where the operation is independently applied to each cropped sub-image feature~$\tilde{F}_{\text{img}}^{i, L-1}$ as follows:

\begin{equation}
    \resizebox{0.9\linewidth}{!}{$
        \tilde{Q}^{i} = \text{Proj}_{Q}(\tilde{F}_{\text{img}}^{i, L-1}), \;
        \tilde{K}^{i} = \text{Proj}_{K}(\tilde{F}_{\text{img}}^{i, L-1}), \;
        \tilde{V}^{i} = \text{Proj}_{V}(\tilde{F}_{\text{img}}^{i, L-  1})
    $}
\end{equation}

\noindent where \textit{C} is the number of crops, \textit{hw} is the number of tokens in each flattened crop feature map, and \textit{d} is the feature dimension. We define a stitching operation $\mathcal{G}(\cdot)$ that stitches these representations into unified global feature spaces:

\begin{equation}
    \resizebox{0.9\linewidth}{!}{$
        Q=\mathcal{G}(\{\tilde{Q}^{i}\}^{C}_{i=1}), K=\mathcal{G}(\{\tilde{K}^{i}\}^{C}_{i=1}),  V=\mathcal{G}(\{\tilde{V}^{i}\}^{C}_{i=1})
    $}
\end{equation}

\noindent where $ Q, K, V \in \mathbb{R^{\text{1}\times \textit{HW}\times \textit{d}}}$ represent the flattened tokens of the entire image obtained by stitching all crops into a single global feature map, with $ HW $ denoting the total number of tokens. The attention is then computed globally as:

\begin{equation}
    \text{StitchAttention} = \text{softmax}\left(\frac{QK^\top}{\tau}\right) V
\end{equation}

\noindent where $\tau$ is a temperature parameter.

By design, Stitch Attention enables attention weights to capture relationships across the entire image rather than being restricted to isolated sub-images, effectively transforming the model from crop-level processing into a unified global attention mechanism. This promotes global semantic coherence and consistent feature interactions, which are crucial for maintaining object continuity and achieving precise segmentation boundaries.

After obtaining the globally coherent feature map from our Stitch Attention module, we follow the attention formulation described in \S.~\ref{subsec:preliminaries}. The resulting feature affinity-based attention map is multiplied with the stitched value features from the CLIP visual encoder, producing the final feature representation. This step effectively transfers the global contextual relationships captured by Stitch Attention into the CLIP feature space, thereby reinforcing semantic consistency across the entire image.

\subsection{Class-Biased Prompt Generation}
\noindent Our Stitch Attention module improves the consistency of visual features across sub-images, producing more coherent segmentation masks. This ensures that regions belonging to the same object are grouped together, significantly enhancing segmentation quality. However, a potential drawback arises: if an incorrect class label is assigned, the enhanced consistency propagates the error over a larger region, amplifying the misclassification. 

To mitigate this, we incorporate the Class-Biased Prompts into the text embedding process. Conventional prompts (e.g., “a photo of \texttt{{\{class\}}}”) provide only generic descriptions, causing ambiguity among similar categories. We instead augment them with about 15 simple, bias-oriented phrases per class, generated by a large language model (e.g., “a large asphalt road without pedestrians” for \texttt{road}). 
Despite their simplicity, these Class-Biased Prompts emphasize distinctive category traits and effectively guide the Stitch Attention module toward more accurate class assignments.

By combining these lightweight prompts with consistent visual features from our Stitch Attention module, segmentation benefits from both stronger regional coherence and more reliable class prediction, achieving a substantial improvement even with minimal prompt construction (details are provided in supplementary).
\begin{table*}[t]
    \centering
    \scriptsize
    \resizebox{\textwidth}{!}{
    \begin{tabular}{lcccccccccc}

    \toprule
    \multirow{2.5}{*}{\textbf{Method}} &  & \multicolumn{3}{c}{\it With a background category} & \multicolumn{5}{c}{\it Without background category} & \multirow{2}{*}{Avg.} \\\cmidrule(lr){3-5}\cmidrule(lr){6-10}
    & & VOC21 & Context60 & Object & VOC20 & City & Context59 & ADE20K & Stuff \\
    \midrule
    
    \rowcolor{lightgray} \multicolumn{11}{c}{OpenAI CLIP ViT-B/16} \vspace{2pt} \\
    CLIP~\cite{clip} & \tiny{ICML'21} & {18.6} & {7.8} & {6.5} & {49.1} & {6.7} & {11.2} & {3.2} & {5.7} & {13.6} \\
    MaskCLIP~\cite{maskclip_tf} & \tiny{ECCV'22} & {43.4} & {23.2} & {20.6} & {74.9} & {24.9} & {26.4} & {11.9} & {16.7} & {30.3} \\
    CLIPtrase~\cite{cliptrase} & \tiny{ECCV'24} & {50.9} & {29.9} & {43.6} & {81.0} & {21.3} & {33.8} & {16.4} & {22.8} & {32.7} \\
    ClearCLIP~\cite{clearclip} & \tiny{ECCV'24} & {51.8} & {32.6} & {33.0} & {80.9} & {30.0} & {35.9} & {16.7} & {23.9} & {38.1} \\
    SCLIP~\cite{sclip} & \tiny{ECCV'24} & {59.1} & {30.4} & {30.5} & {80.4} & {32.2} &  {34.2} & {16.1} & {22.4} & {38.2} \\
    NACLIP~\cite{naclip} & \tiny{WACV'25} & {58.9} & {32.2} & {33.2} & {79.7} & {35.5} & {35.2} & {17.4} & {23.3} & {39.4} \\
    ResCLIP~\cite{resclip} & \tiny{CVPR'25} & {61.1} & {33.5} & {35.0} & {86.0} & {35.9} & {36.8} & {18.0} & {24.7} & {41.4} \\
    ProxyCLIP~\cite{proxyclip} & \tiny{ECCV'24} & {61.3} & {35.3} & {37.5} & {80.3} & {38.1} & {39.1} & {20.2} & {26.5} & {42.3} \\
    SC-CLIP~\cite{sc_clip} & \tiny{Arxiv'24} & {64.6} & {36.8} & {37.7} & {84.3} & {41.0 } & {40.1} & {20.1} & {26.6} & {43.9} \\
    SFP~\cite{sfp} & \tiny{ICCV'25} & {62.9} & {37.2} & {37.9} & {84.5} & {41.1} & {39.9} & {20.8} & {26.4} & {44.0} \\
    CASS~\cite{cass} & \tiny{CVPR'25} & {65.8} & {36.7} & {37.8} & {87.8} & {39.4} & {40.2} & {20.4} & {26.7} & {44.4} \\
    Trident~\cite{trident} & \tiny{ICCV'25} & {67.1} & {38.6} & \second{41.1} & {84.5} & {42.9} & {42.2} & {21.9} & {28.3} & {45.8} \\

    CorrCLIP~\cite{corrclip} & \tiny{ICCV'25} & \second{72.2} & \second{41.6} & {40.7} & \second{88.7} & \second{44.6} & \second{47.1} & \second{23.7} & \second{30.7} & \second{48.7} \\    
    
    \gray{~~~{\raisebox{0.5ex}{\large$\llcorner$}} w/o post-processing} & & \gray{69.2} & \gray{40.0} & \gray{39.8} & \gray{87.0} & \gray{41.6} & \gray{44.9} & \gray{22.4} & \gray{29.6} & \gray{46.8} \\
    
    \rowcolor{gray!10}
    \textbf{OV-Stitcher} & \tiny{Ours} & \bf{75.7} & \bf{43.9} & \bf{42.6} & \bf{89.8} & \bf{48.1} & \bf{48.8} & \bf{24.7} & \bf{31.8} & \bf{50.7} \\
        \rowcolor{gray!10}
        
    \gray{~~~{\raisebox{0.5ex}{\large$\llcorner$}} w/o post-processing} & & \gray{73.1} & \gray{42.4} & \gray{41.4} & \gray{87.6} & \gray{45.4} & \gray{47.1} & \gray{23.6} & \gray{30.7} & \gray{48.9} \\
    \midrule

    \rowcolor{lightgray} \multicolumn{11}{c}{OpenAI CLIP ViT-L/14} \vspace{2pt}  \\
    CLIP~\cite{clip} & \tiny{ICML'21} & {8.2} & {4.1} & {2.7} & {15.6} & {4.4}  & {2.5} & {1.7} & {2.4} & {5.2} \\
    MaskCLIP~\cite{maskclip_tf} & \tiny{ECCV'22} & {23.3} & {11.7} & {7.2} & {29.4} & {12.4} & {11.5} & {7.2} & {8.8} & {13.9} \\
    ResCLIP~\cite{resclip} & \tiny{CVPR'25} & {54.1} & {30.9} & {32.5} & {85.5} & {33.7} & {34.5} & {18.2} & {23.4} & {39.1} \\
    ProxyCLIP~\cite{proxyclip} & \tiny{ECCV'24} & {60.6} & {34.5} & {39.2} & {83.2} & {40.1} & {37.7} & {22.6} & {25.6} & {43.0} \\
    SC-CLIP~\cite{sc_clip} & \tiny{Arxiv'24} & {65.0} & {36.9} & {40.5} & {88.3} & {41.3 } & {40.6} & {21.7} & {26.9} & {45.2}\\
    Trident~\cite{trident} & \tiny{ICCV'25} & {62.6} & {37.3} & {40.5} & {85.5} & {43.0} & {40.9} & \second{24.0} & {27.1} & {45.1} \\
    CorrCLIP~\cite{corrclip} & \tiny{ICCV'25} & \second{71.8} & \second{42.2} & \second{46.2} & \bf{91.2} & \second{47.9} & \second{47.2} & \bf{27.7} & \second{31.0} & \second{50.6} \\

    \rowcolor{gray!10}
    \textbf{OV-Stitcher} & \tiny{Ours} & \bf{74.0} & \bf{43.4} & \textbf{46.5} & \second{90.2} & \bf{50.6} & \bf{48.6} & \bf{27.7} & \bf{31.6} & \bf{51.6} \\
    \midrule

    \rowcolor{lightgray} \multicolumn{11}{c}{MetaCLIP ViT-B/16} \vspace{2pt}  \\
    ProxyCLIP~\cite{proxyclip} & \tiny{ECCV'24} & {63.3} & {37.5} & {38.4} & {81.0} & {39.9} & {40.8} & {22.5} & {28.1} & {43.9} \\
    Trident~\cite{trident} & \tiny{ICCV'25} & {68.4} & {39.9} & {41.7} & {85.4} & {43.6} & {46.1} & {23.7} & {29.8} & {47.4} \\
    CorrCLIP~\cite{corrclip} & \tiny{ICCV'25} & \second{74.8} & \second{44.2} & \second{43.7} & \bf{88.8} & \second{49.4} & \second{48.8} & \second{26.9} & \second{31.6} & \second{51.0} \\

    \rowcolor{gray!10}
    \textbf{OV-Stitcher} & \tiny{Ours} & \bf{76.4} & \bf{43.9} & \bf{44.6} & \second{88.7} & \bf{52.3} & \bf{49.1} & \bf{27.8} & \bf{32.1} & \bf{51.9} \\
    \bottomrule

    \end{tabular}
    }
    \caption{ \textbf{Quantitative Comparison of Prior Open Vocabulary Segmentation Works.} The highest-performing result is highlighted in \textbf{bold}, and the second highest in \second{underline} for clarity. The \textit{``w/o post-processing"} rows show the performance without the post-processing step, where each SAM-generated mask is assigned the label corresponding to the most frequent raw logit prediction within that mask.}
\label{tab:main_result}
\vspace{-3mm}
\end{table*}

\section{Experiments}
\label{sec:experiments}
\begin{figure*}[t!]
    \centering
    \includegraphics[width=\textwidth]{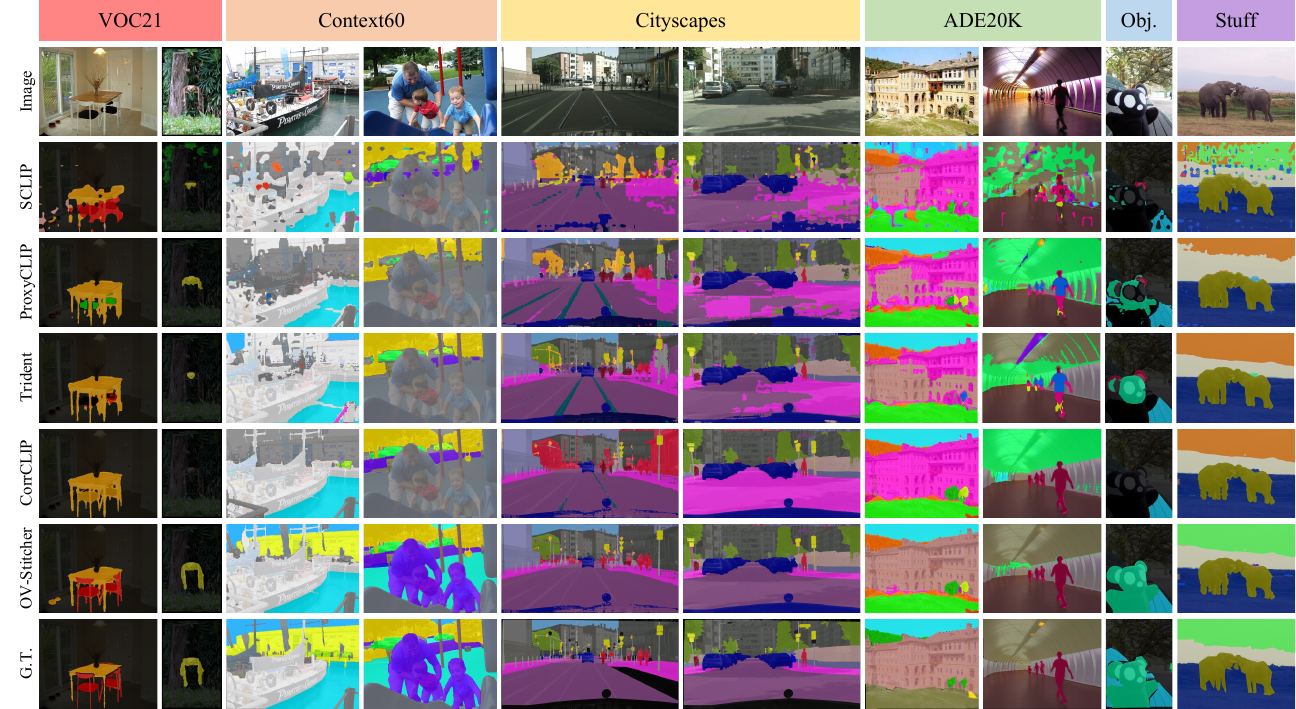}
    \caption{\textbf{Qualitative comparison with previous training-free open vocabulary segmentation methods.}}
    \label{fig:qualitative_result}
    \vspace{-3mm}
\end{figure*}

\subsection{Experimental setup}

\textbf{Dataset.} We evaluate OV-Stitcher on eight open-vocabulary semantic segmentation benchmarks derived from six widely used datasets: \textbf{PASCAL VOC 2012}~\cite{pascal_voc}, \textbf{PASCAL Context}~\cite{pascal_context}, \textbf{COCO Object}~\cite{coco_obj}, \textbf{COCO Stuff}~\cite{coco_stuff}, \textbf{Cityscapes}~\cite{cityscapes}, and \textbf{ADE20K}~\cite{ade20k}.
For PASCAL VOC and Context, we follow two settings depending on whether background categories are included—VOC20/VOC21 and Context59/Context60—resulting in eight benchmarks in total.
The design of Stitch Attention supports high-resolution inputs, allowing flexible image sizes across datasets. The shorter side is set to 448 pixels for all datasets except Cityscapes (560 pixels). Sliding-window inference uses 336×336 crops with a stride of 112 pixels, while Cityscapes uses 224×224 crops and COCO Stuff a 224-pixel stride.

\noindent
\textbf{Baselines and Comparison Methods.}
We conduct experiments using OpenAI CLIP ~\cite{clip} with ViT-B/16 and ViT-L/14 backbones as the primary vision–language models.
For the feature extractor, we use DINO~\cite{dino} ViT-B/8 to obtain spatial representations.
The Class-Biased Prompts Generator is implemented using LLaMA3 8B~\cite{llama3}, where class-specific text embeddings are precomputed and utilized during inference. To generate class-biased prompts, we prompt LLaMA3 to produce 15 general descriptive sentences for each class, capturing visual attributes such as shape, texture, and other salient characteristics.
\vspace{-1px}

Moreover, we compare OV-Stitcher against a broad set of recent TF-OVSS approaches, including CLIP~\cite{clip}, MaskCLIP~\cite{maskclip_tf}, ClearCLIP~\cite{clearclip}, SCLIP~\cite{sclip}, NaCLIP~\cite{naclip}, ResCLIP~\cite{resclip}, SC-CLIP~\cite{sc_clip}, ProxyCLIP~\cite{proxyclip}, SFP~\cite{sfp}, CASS~\cite{cass}, Trident~\cite{trident}, and CorrCLIP~\cite{corrclip}.
Since one of the compared methods reports results with MetaCLIP~\cite{metaclip}, we re-evaluate that method using OpenAI CLIP for a fair comparison.
We also evaluate several baselines, as well as our method, using MetaCLIP ViT-B/16 to ensure consistency across settings.

Our method follows the CorrCLIP framework, utilizing masks from SAM2~\cite{sam2} with MAE~\cite{mae} pretrained Hiera-L~\cite{hiera, hiera_pos} to mask the attention map and to perform post-processing.
We adopt the reference implementation of CorrCLIP as our baseline setup, which allows us to evaluate OV-Stitcher across a variety of experiments in comparison to CorrCLIP, providing an intuitive view of our approach’s effectiveness.
We compare results using mean Intersection over Union (mIoU).
All experiments are implemented using the MMSegmentation~\cite{mmseg, mmengine} framework.

\subsection{Main Results}
\textbf{Quantitative results.} The results, summarized in Tab.~\ref{tab:main_result}, clearly demonstrate the advantage of OV-Stitcher.
With the ViT-B/16 backbone, OV-Stitcher achieves state-of-the-art performance on every benchmark, surpassing the previously best-performing model by about 2.0\% mIoU on average.
When evaluated with ViT-L/14, OV-Stitcher continues to deliver top performance on most benchmarks and attains the highest average score among all methods.
As shown in the lower part of Tab.~\ref{tab:main_result} for MetaCLIP, OV-Stitcher again achieves the highest performance across all datasets, and the average score further improves by 1.2\% in mIoU compared to the OpenAI CLIP results, reflecting the benefit of stronger visual representations.

Overall, a notable observation is that OV-Stitcher achieves particularly large gains on the Cityscapes dataset, outperforming previous methods by a substantial margin with increases of 3.5\%, 2.7\%, and 2.9\% in averaged mIoU across the three variants.
Since Cityscapes contains a relatively large number of cropped sub-images per sample, the Stitch Attention mechanism can more effectively integrate cross-crop contextual cues, leading to more coherent and consistent predictions.

Taken together, the results indicate that our proposed stitching mechanism generalizes effectively across different backbones, including larger variants, and that stitching local and global contexts is highly effective in alleviating the spatial fragmentation problem inherent in prior training-free open-vocabulary segmentation frameworks.

\noindent \textbf{Qualitative results.} As shown in Fig.~\ref{fig:qualitative_result}, OV-Stitcher produces segmentation maps with improved spatial coherence and more accurate class alignment compared to previous training-free approaches. 

While CorrCLIP may appear to show a comparable level of feature consistency across regions, this perceived coherence mainly results from the post-processing step of the segmentation map correction module, which refines each mask from SAM2 by assigning the most frequent class label within it.
To highlight the true contribution of Stitch Attention itself, we therefore present segmentation results obtained directly from the raw logits, without any post-processing.
A detailed discussion of segmentation results obtained without post-processing is provided in \S.~\ref{subsec:ablation}.

\subsection{Ablation Study}
\label{subsec:ablation}
Since our framework is built upon the reference implementation of CorrCLIP, it naturally serves as our baseline. This setup allows us to conduct a variety of comparative experiments between OV-Stitcher and CorrCLIP, providing an intuitive understanding of the effectiveness of each proposed component.

\noindent \textbf{Effectiveness of Each Component.} We conducted an ablation study, summarized in Tab.~\ref{tab:abl_components}, to assess the contributions of Stitch Attention (StitchAttn) and Class Biased Prompts (CBP).
The baseline model without StitchAttn or CBP achieves reasonable segmentation performance. Introducing CBP alone, which augments the standard ImageNet templates with CBP to reduce ambiguity in text queries, consistently improves the model’s ability to distinguish between classes.
Applying StitchAttn alone also enhances performance, demonstrating that stitching local and global contexts contributes to greater semantic consistency in segmentation predictions.
When both StitchAttn and CBP are combined, the model achieves the best results, confirming that the two components are complementary: StitchAttn improves spatial and semantic coherence, while CBP reduces ambiguity in text queries. Together, they lead to the most accurate, consistent, and coherent segmentations, validating the design choices of OV-Stitcher.

\begin{figure}[t]
    \centering
    \includegraphics[width=\columnwidth]{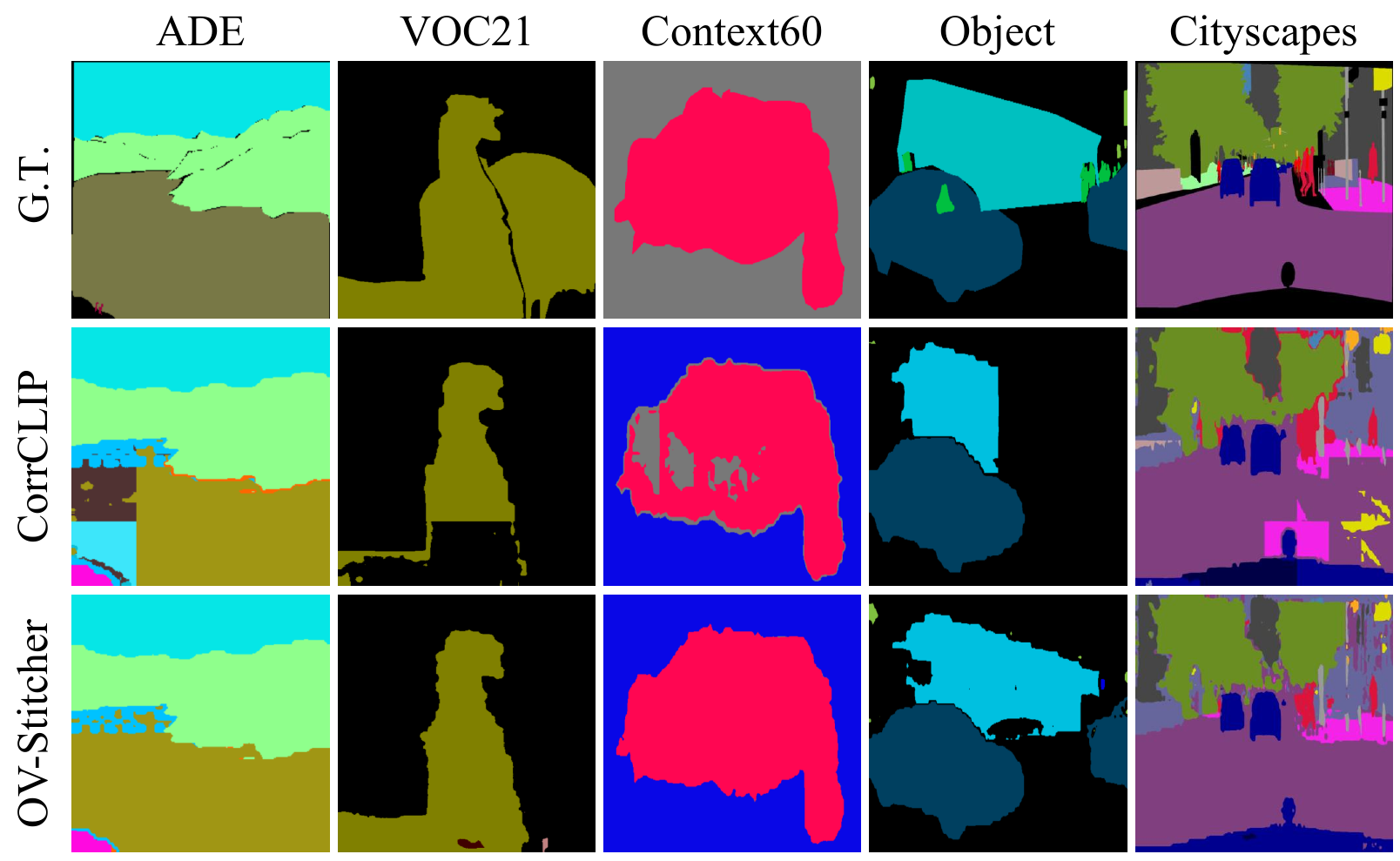}
    \caption{\textbf{Qualitative comparison without post-processing.}}
    \label{fig:qualitative-no-mc}
\end{figure}

\begin{table}[t]
\centering
\resizebox{\columnwidth}{!}{
\begin{tabular}{cc|ccccccc}
    \toprule
    StitchAttn & CBP & V21 & C60 & Obj. & Stf. & City & ADE \\ \hline
    {} & {} & {72.2} & {41.6} & {40.7} & {30.7} & {44.6} & {23.7}     \\
    {} & {\checkmark} & {73.2} & {42.7} & {41.6} & {31.0} & {45.9} & {24.3}     \\
    {\checkmark} &  & {74.7} & {42.7} & {42.3} & {31.2} & {46.7} & {23.9}     \\
    {\checkmark} & {\checkmark} &  \textbf{75.7} &  \textbf{43.7} &  \textbf{42.7} &  \textbf{31.8} &  \textbf{48.1} &  \textbf{24.7}     \\
    \bottomrule
\end{tabular}
}
\caption{\textbf{Ablation study evaluating the impact of each component in our proposed method.}}
\label{tab:abl_components}
\vspace{-4mm}
\end{table}

\noindent \textbf{Evaluation Without Post-processing.} To better assess OV-Stitcher’s effectiveness without post-processing, we evaluate the predictions obtained directly from the raw logits. As shown in Tab.~\ref{tab:main_result}, \textit{``w/o post-processing"} rows, OV-Stitcher outperforms CorrCLIP even without post-processing, demonstrating that the proposed approach produces strong and accurate predictions at the logit level.
Fig.~\ref{fig:qualitative-no-mc} illustrates that qualitative results further highlight how OV-Stitcher reduces fragmentation within regions sharing the same semantic meaning, yielding more coherent and consistent segmentation maps. While post-processing in the main experiments smooths differences, this ablation clearly shows that the model itself, through the stitching mechanism, achieves better semantic consistency across the image.

\begin{figure}[t]
    \centering
    \includegraphics[width=\columnwidth]{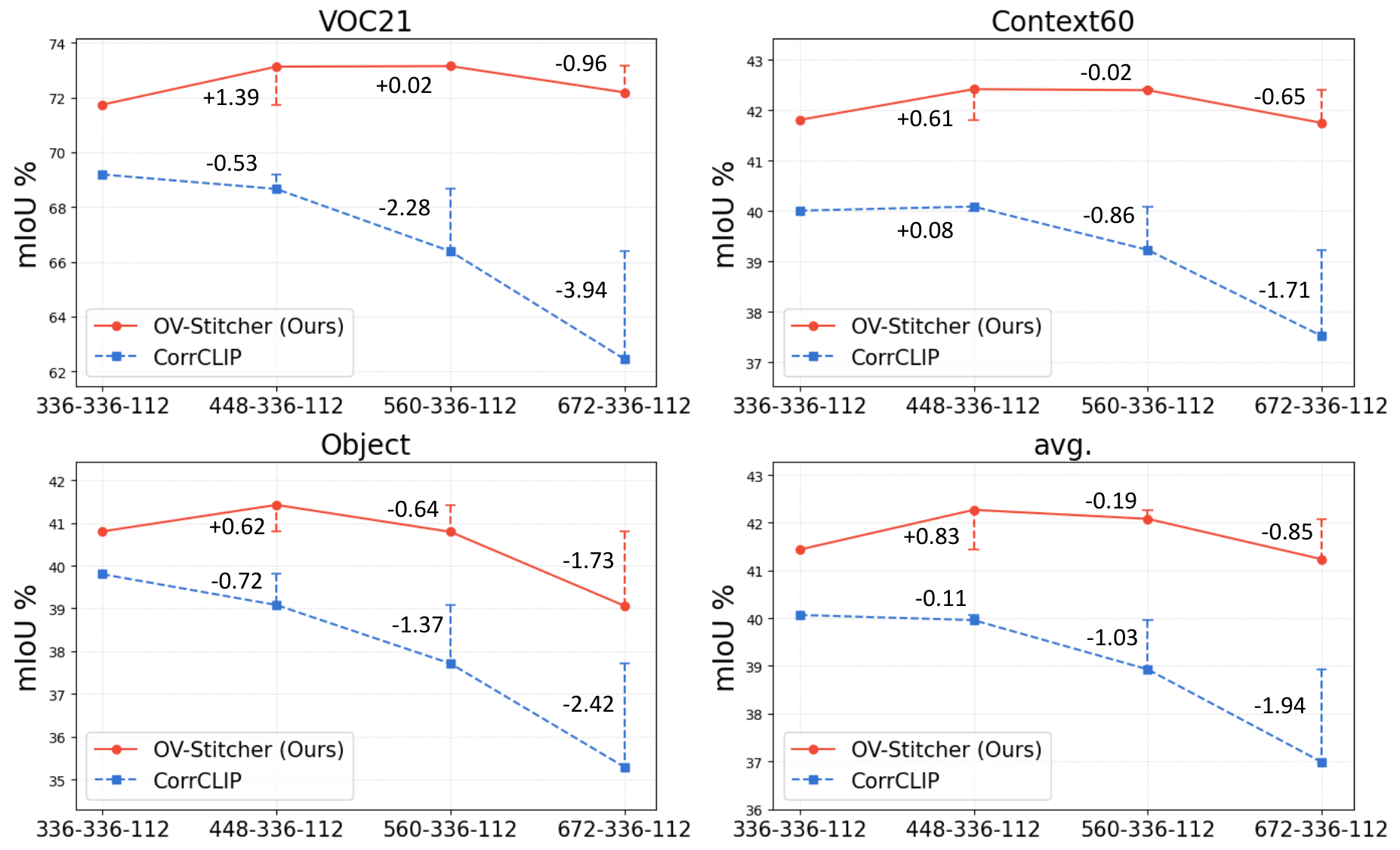}
    \caption{\textbf{Ablation on resolution robustness.}
    Post-processing is excluded to clearly show the effect of the proposed framework. The x-axis represents the settings in the format \textbf{shorter side} – \textbf{window size} – \textbf{stride}.}
    \label{fig:resolution-graph}
    \vspace{-0.2cm}
\end{figure}

\noindent \textbf{Performance under Varying Resolutions.}
High-resolution inputs often lead to a loss of consistency in segmentation when each crop is processed independently, as in previous approaches. Since our stitching mechanism allows all sub-images to attend to each other during feature aggregation, it is expected to maintain stronger robustness when processing images with a large number of crops at high resolutions. To verify this, we conduct an ablation study comparing OV-Stitcher with the baseline method CorrCLIP under identical settings. As shown in Fig.~\ref{fig:resolution-graph}, while performance of OV-Stitcher remains stable or even slightly improves as input resolution increases, performance of CorrCLIP drops significantly, demonstrating the effectiveness of our stitching mechanism in maintaining robust segmentation across high-resolution inputs (results on other datasets are provided in supplementary). 

\noindent \textbf{Effectiveness of Stitch Attention.} Stitch Attention facilitates the transfer of spatial information from Vision Foundation Model (VFM) features, such as those from DINO, to a CLIP-based representation.
In the same vein, this mechanism can be applied to ProxyCLIP, a baseline method leveraging VFM-derived spatial features. As shown in Tab.~\ref{tab:abl_proxy}, we apply Stitch Attention to ProxyCLIP and observe consistent improvements in segmentation performance, demonstrating that the approach effectively enhances spatial coherence and can generalize beyond a single framework.

\begin{table}[t]
\centering
\resizebox{\columnwidth}{!}{
\begin{tabular}{c|ccccccc}
    \toprule
     ProxyCLIP& V21 & C60 & Obj. & Stf. & City & ADE \\ \hline
    {X} & {61.3} & {35.3} & {37.5} & {26.5} & {38.1} & {20.2} \\
    {\checkmark} & \textbf{62.9} & \textbf{36.3} & \textbf{38.1} & \textbf{26.7} & \textbf{39.8} & \textbf{20.9} \\
    \bottomrule
\end{tabular}
}
\caption{\textbf{Effectiveness of Stitch Attention on Other Method.} “X” indicates the original ProxyCLIP; “\checkmark” indicates ProxyCLIP with Stitch Attention.}
\label{tab:abl_proxy}
\vspace{-4mm}
\end{table}

\section{Conclusion}
\vspace{-1mm}
\label{sec:conclusion}
In this work, we introduced OV-Stitcher, a framework that enhances training-free open-vocabulary segmentation by integrating global context across sub-images through the Stitch Attention mechanism.
By allowing cross-crop feature interactions, OV-Stitcher mitigates the spatial fragmentation inherent in prior training-free approaches, maintaining semantic coherence and accurate object boundaries even at high resolutions.
Additionally, the incorporation of Class-Biased Prompts further reduces ambiguity in text embeddings, improving class-level alignment.
By combining these design choices, our method achieves notable improvements in segmentation performance, leading to superior results across a diverse set of benchmarks.

\section{Acknowledgments}
\label{sec:acknowledgement}
This work was supported by the National Research Foundation (NRF) grant funded by the Korea government (MSIT) [RS-2025-00562400] and [RS-2022-NR068754].

{
    \small
    \bibliographystyle{ieeenat_fullname}
    \bibliography{main}
}

\clearpage
\maketitlesupplementary

\setcounter{section}{0}%
\renewcommand{\thesection}{\Alph{section}}%
\renewcommand{\thesubsection}{\thesection\arabic{subsection}}%

\begin{figure*}[!t]
    \centering
    \includegraphics[width=0.95\textwidth]{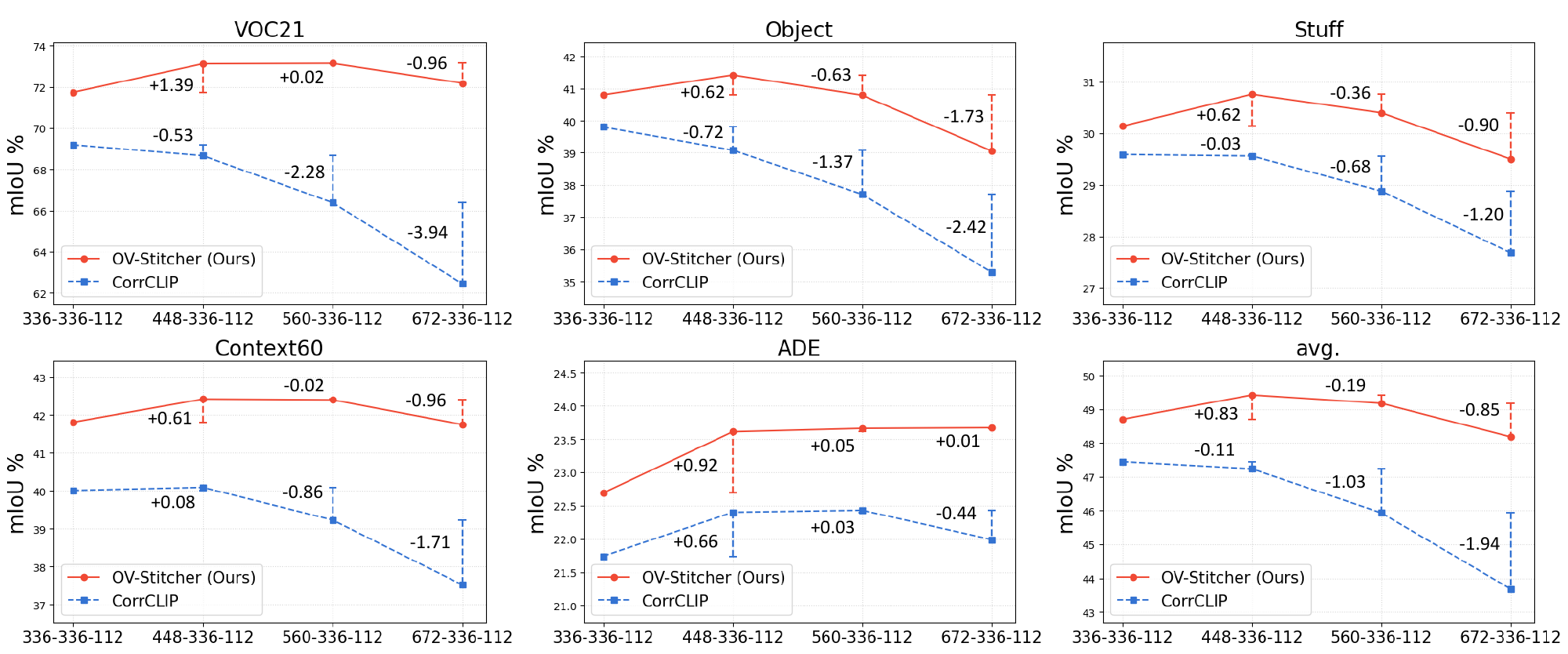}
    \caption{\textbf{Ablation on resolution robustness.} Post-processing is excluded to clearly show the effect of the proposed framework. The x-axis represents the settings in the format \textbf{shorter side} – \textbf{window size} – \textbf{stride}.}
    \label{fig:sup-res}
    \vspace{-3mm}
\end{figure*}

\section{Additional Results on Varying Resolutions.}
\label{sup-A:resolution}
To complement the results presented in the main paper, we provide additional evaluations on the same resolution settings using both graphical summaries (Fig.~\ref{fig:sup-res}) and numerical results (Tab.~\ref{tab:sup-res}). In addition to VOC21~\cite{pascal_voc}, Context60~\cite{pascal_context}, and COCO Object~\cite{coco_obj}, this section includes results on ADE20K~\cite{ade20k} and COCO Stuff~\cite{coco_stuff}.
Following the same setup, we fix the window size and stride to 336 and 112, respectively, and vary the shorter side from 336 to 448, 560, and 672, increasing the number of sub-image crops. Under identical conditions, we compare OV-Stitcher with CorrCLIP~\cite{corrclip} across all datasets.

As shown in Tab.~\ref{tab:sup-res} and Fig.~\ref{fig:sup-res}, OV-Stitcher consistently shows smaller drops in performance compared to CorrCLIP as resolution increases. While CorrCLIP generally achieves its best results at the lowest resolution across datasets, OV-Stitcher attains its peak performance at higher resolutions, demonstrating the effectiveness of the stitching mechanism in leveraging high-resolution inputs.

\section{Computational Analysis.}
\label{sup-B:computational_cost}
While computational efficiency is not the primary focus of our method, Stitch Attention introduces full token-to-token interaction across sub-images, which makes it necessary to examine how inference cost scales with input size.
Our proposed Stitch Attention enables attention over all tokens across sub-images, allowing the model to capture global context effectively.
However, this also introduces a dependency of inference cost on both input resolution and the total number of tokens, which varies per image.
Therefore, we evaluate the computational cost at fixed resolutions of 336×336, 448×448, and 560×560. For these experiments, we adopt a sliding-window configuration with a window size of 336 and a stride of 112, allowing us to measure how inference cost increases sequentially with input size.

As expected, higher resolutions lead to a larger number of tokens and consequently higher computation, as shown in Table~\ref{tab:sup-cost}.
In particular, increasing the resolution results in more crops being processed, which further amplifies the computational load.
Nonetheless, our method effectively leverages higher-resolution inputs to yield improved segmentation performance, as reported in Table~\ref{tab:sup-res}, making this additional computation a reasonable trade-off and highlighting the advantage of maintaining global interactions even at large input scales.

\begin{table}[htbp]
\setlength{\tabcolsep}{4pt}
\renewcommand{\arraystretch}{1.1}
\resizebox{\columnwidth}{!}{
\begin{tabular}{c|rrrrr|r}
    \toprule
    \bf{Resolution} & \bf{V21} & \bf{C60} & \bf{Obj.} & \bf{Stf.} & \bf{ADE} & \bf{avg.}\\
    \bottomrule
    \rowcolor{gray!20} \multicolumn{7}{c}{\textbf{OV-Stitcher}} \\
    \addlinespace[0.5ex]
    336$^{(336\text{–}112)}$ & {71.74} & {41.81} & {40.80} & {30.14} & {22.69} & {41.43}\\
    448$^{(336\text{–}112)}$ & {73.13} & \bf{42.42} & \bf{41.42} & \bf{30.76} & {23.61} & \bf{42.27}\\
    560$^{(336\text{–}112)}$ & \bf{73.15} & {42.40} & {40.79} & {30.40} & {23.66} & {42.08}\\
    672$^{(336\text{–}112)}$ & {72.19} & {41.75} & {39.06} & {29.50} & \bf{23.67} & {41.23}\\
    \bottomrule
    \rowcolor{gray!20} \multicolumn{7}{c}{\textbf{CorrCLIP}} \\
    \addlinespace[0.5ex]
    336$^{(336\text{–}112)}$ & \bf{69.19} & {40.01} & \bf{39.80} & \bf{29.59} & {21.74} & \bf{40.07}\\
    448$^{(336\text{–}112)}$ & {68.66} & \bf{40.09} & {39.08} & {29.56} & {22.40} & {39.96}\\
    560$^{(336\text{–}112)}$ & {66.38} & {39.23} & {37.71} & {28.88} & \bf{22.43} & {38.93}\\
    672$^{(336\text{–}112)}$ & {62.44} & {37.52} & {35.29} & {27.68} & {21.99} & {36.98}\\
    \bottomrule
\end{tabular}
}
\captionof{table}{
    \textbf{Ablation on resolution robustness.}
    Comparison between CorrCLIP and WeaveCLIP under varying input resolutions without post-processing to clearly show the effect of the proposed framework.
    Each resolution is denoted as \textbf{shorter side}$^{(\textbf{window size - stride})}$.
}
\label{tab:sup-res}
\vspace{-2mm}

\end{table}

\begin{table}[htbp]
\setlength{\tabcolsep}{4pt}
\renewcommand{\arraystretch}{1.1}
\centering
    \resizebox{\columnwidth}{!}{
    \begin{tabular}{c|c|ccc}
        \toprule
        \textbf{Input Res.} & \bf{\# Crops} & \textbf{\# Params. (M)} & \textbf{Mem. (MB)} & \textbf{Thru. (img/sec)} \\
        \bottomrule
        \rowcolor{gray!20} \multicolumn{5}{c}{\textbf{Precomputed Masks}} \\
        \addlinespace[0.5ex]
        336 $\times$ 336 & 1 & 235 & 1435 & 6.98 \\
        448 $\times$ 448 & 4 & 235 & 1450 & 4.72 \\
        560 $\times$ 560 & 9 & 235 & 2040 & 3.12 \\
        672 $\times$ 672 & 16 & 235 & 3198 & 2.12 \\
        \bottomrule
        \rowcolor{gray!20} \multicolumn{5}{c}{\textbf{Masks Generated On-the-Fly}} \\
        \addlinespace[0.5ex]
        336 $\times$ 336 & 1 & 458 & 2627 & 1.58 \\
        448 $\times$ 448 & 4 & 458 & 2651 & 1.47 \\
        560 $\times$ 560 & 9 & 458 & 2691 & 1.25 \\
        672 $\times$ 672 & 16 & 458 & 3717 & 1.03 \\
        \bottomrule
    \end{tabular}
    }
\caption{\textbf{Computational costs on RTX 4090 with FP16.} 
We separate cases where SAM2 masks for highlighting the attention map and post-processing are precomputed from those where they are generated on-the-fly.}
\label{tab:sup-cost}
\vspace{-2mm}
\end{table}

\begin{table}[htbp]
\centering
\resizebox{\columnwidth}{!}{
    \begin{tabular}{c|cc|cc}
        \hline
        \multirow{2}{*}{Input Res.} & \multicolumn{2}{c|}{\textbf{Naive Ver.} Stitch Attention} & \multicolumn{2}{c}{\textbf{Flash Ver.} Stitch Attention} \\
        & Latency (ms) & Memory (MB) & Latency (ms) & Memory (MB) \\
        \hline
        
        \multicolumn{1}{c|}{336$\times$336} &
        0.25 & 293 &
        0.21~\tiny{($\blue{16.0\% \downarrow}$)} & 220 \tiny{($\blue{24.9\% \downarrow}$)}  \\
        
        \multicolumn{1}{c|}{448$\times$448} &
        0.72 & 454 &
        0.44~\tiny{($\blue{38.9\% \downarrow}$)} & 241 \tiny{($\blue{46.9\% \downarrow}$)}  \\
        
        \multicolumn{1}{c|}{560$\times$560} &
        1.65 & 799 &
        0.81 \tiny{($\blue{50.9\% \downarrow}$)}  & 273 \tiny{($\blue{65.8\% \downarrow}$)}  \\
        
        \multicolumn{1}{c|}{672$\times$672} &
        3.08 & 1423 &
        1.48 \tiny{($\blue{52.0\% \downarrow}$)}  & 317 \tiny{($\blue{77.7\% \downarrow}$)}  \\
        \hline
    \end{tabular}
}
\caption{\textbf{Computational costs on RTX 4090 with FP16.} Latency and peak CUDA memory of StitchAttention with naive attention and Flash Attention at different resolutions.}
\label{tab_reb:flash_attn}
\vspace{-5mm}
\end{table}

Moreover, to examine the practical feasibility of our approach, we apply Flash Attention\cite{flashattn2} to the Stitch Attention module by replacing the attention computation, while keeping the rest of the framework unchanged. As shown in Table~\ref{tab_reb:flash_attn}, this consistently reduces both latency and peak memory across all resolutions, with larger gains at higher resolutions (e.g., 52.0\% latency and 77.7\% memory reduction at 672$\times$672). These results indicate that the additional cost introduced by global token interactions can be effectively mitigated using standard efficient attention implementations, supporting the practical applicability of our method.

To generate Class-Biased Prompts, we employ a Large Language Model (LLM). Empirically, generating 15 descriptions per class required an average of approximately 5 seconds, which can pose a computational burden. However, this computational burden can be alleviated either by reducing the number of descriptions per class or by precomputing prompts—even if this slightly deviates from the fully open-vocabulary scope of OVSS—for a massive set of classes with extensive vocabularies (e.g., ImageNet-21K~\cite{in21k}, Open Images Dataset~\cite{oid}).
\vspace{-2mm}

\section{Evaluation on Diverse CLIP Variants.}
\label{sup-C: CLIP}
We additionally evaluate our method using various CLIP backbones, including OpenCLIP~\cite{laionclip}, MetaCLIP~\cite{metaclip}, and DFNCLIP~\cite{dfnclip} with both ViT-B/16 and ViT-L/14. As shown in Tab.~\ref{tab:sup-clip}, results vary noticeably across CLIP variants: MetaCLIP still delivers the highest overall performance, but its improvement is less pronounced compared to the substantial jump observed from the Base models, whereas OpenCLIP and DFNCLIP exhibit more moderate gains across datasets.

Interestingly, higher zero-shot classification accuracy does not necessarily translate into stronger segmentation performance, likely because segmentation relies more on spatial detail and local region consistency than on the global semantic discrimination emphasized during CLIP pretraining. Since our method directly leverages CLIP’s value features, it would be valuable for future work to explore how these value representations could retain or enhance spatial information, potentially improving segmentation robustness across diverse datasets.
\begin{table}[htbp]
\centering
\setlength{\tabcolsep}{4pt}
\renewcommand{\arraystretch}{1.1}
\resizebox{\columnwidth}{!}{
    \begin{tabular}{c|c|c|ccccccc}
    \toprule
    \bf{Type} & \bf{Size} & \bf{Acc.} & \bf{V21} & \bf{C60} & \bf{Obj.} & \bf{Stf.} & \bf{City} & \bf{ADE} \\
    \midrule
    OpenCLIP  & \multirow{3}{*}{ViT-B/16} & 70.2\% & 72.92 & 40.54 & 42.27 & 31.20 & 51.07 & 28.01 \\
    MetaCLIP  & & 72.1\% &\bf{76.37} & \bf{43.92} & \bf{44.59} & \bf{32.10} & \bf{52.26} & \bf{27.80} \\
    DFNCLIP & & 76.2\% & 72.26 & 41.82 & 43.93 & 32.16 & 52.09 & 27.78 \\
    \midrule
    OpenCLIP  & \multirow{3}{*}{ViT-L/14} & 75.3\% & 73.10 & 41.55 & 42.16 & 31.02 & 52.82 & 28.10 \\
    MetaCLIP & & 79.2\% & \bf{76.47} & \bf{45.50} & \bf{49.75} & \bf{34.47} & \bf{53.01} & \bf{30.59} \\
    DFNCLIP & & 81.4\% & 74.58 & 43.39 & 43.13 & 33.77 & 51.77 & 28.63 \\
    \bottomrule
    \end{tabular}
    }
    \caption{\textbf{Comparison of different CLIP variants used as vision–language backbones.} Acc. denotes the zero-shot classification accuracy of each CLIP model on ImageNet-1K~\cite{in1k}}
    \label{tab:sup-clip}

\end{table}

\begin{table}[t]
\setlength{\tabcolsep}{4pt}
\renewcommand{\arraystretch}{1.1}
\centering
    \resizebox{\columnwidth}{!}{
    \begin{tabular}{c|c|cccccc}
        \toprule
        \textbf{Type} & \textbf{Size} & \textbf{V21} & \textbf{C60} & \textbf{Obj.} & \textbf{Stf.} & \textbf{City} & \textbf{ADE} \\
        \midrule
        \multirow{3}{*}{DINO}
        & ViT-S/8 & \bf{75.84} & \second{43.72} & \bf{42.63} & \bf{31.86} & \second{47.05} & \second{24.61} \\
        & ViT-B/8  & \second{75.72} & \bf{43.85} & \second{42.55} & \second{31.83} & \bf{48.06} & \bf{24.72} \\
        & ViT-B/16 & 75.47 & 43.68 & 41.77 & 31.69 & 46.58 & 24.26 \\
        \midrule
        \multirow{2}{*}{DINOv2}
        & ViT-B/14  & 75.42 & 43.41 & 42.14 & 31.53 & 45.23 & 24.31 \\
        & ViT-L/14 & 75.28 & 43.21 & 42.54 & 31.45 & 44.14 & 24.36 \\
        \bottomrule
    \end{tabular}
    }
\caption{\textbf{Evaluation of various feature extractors.}}
\label{tab:sup-extractor}

\vspace{-5mm}
\end{table}

\section{Evaluation on Diverse Featrue Extractor.}
\label{sup-D: DINO}
We evaluate our approach using a range of self-supervised feature extractors, including DINO~\cite{dino} and DINOv2~\cite{dinov2} variants with different backbone sizes and patch resolutions. As shown in Tab.~\ref{tab:sup-extractor}, models with smaller patch sizes (e.g., ViT-S/8 and ViT-B/8) consistently deliver higher segmentation quality than architectures with larger patch sizes, even when those models are stronger or larger overall. This highlights the importance of preserving detailed spatial information, which is more naturally retained with finer patch granularity.

Although DINO-B/8 slightly outperforms its smaller counterpart, DINO-S/8, the gap remains relatively modest.
Considering the increased computational cost of larger models, this suggests a practical trade-off: lightweight models with small patch sizes—such as DINO-S/8—can offer competitive segmentation performance while improving inference speed and reducing memory consumption.
 
\section{Class-Biased Prompts Construction.}
\label{sup-E: CBP}
To obtain class-biased prompts, we used an LLM to generate fine-grained visual descriptions tailored to each category. In addition to conventional ImageNet-style templates (e.g., “a photo of \texttt{{\{class\}}}”), we designed a set of instructions that guide the model to produce diverse, visually grounded sentences highlighting typical and distinctive attributes of the target class.

The LLM was instructed to produce 15 concise descriptions (5–15 words) for each class, highlighting features such as shape, surface appearance, material, structural components, and typical visual contexts, and to describe each class from multiple visual perspectives—for example, by emphasizing form, texture, surrounding environment, or characteristic parts.

A simplified version of the instruction used is:

\begin{itemize}
    \item\textit{“Generate 15 concise visual descriptions of a }\texttt{{\{class\}}}\textit{, focusing only on typical, observable features such as shape, material, or context.”}
\end{itemize}

\begin{figure*}
    \centering
    \includegraphics[width=\textwidth]{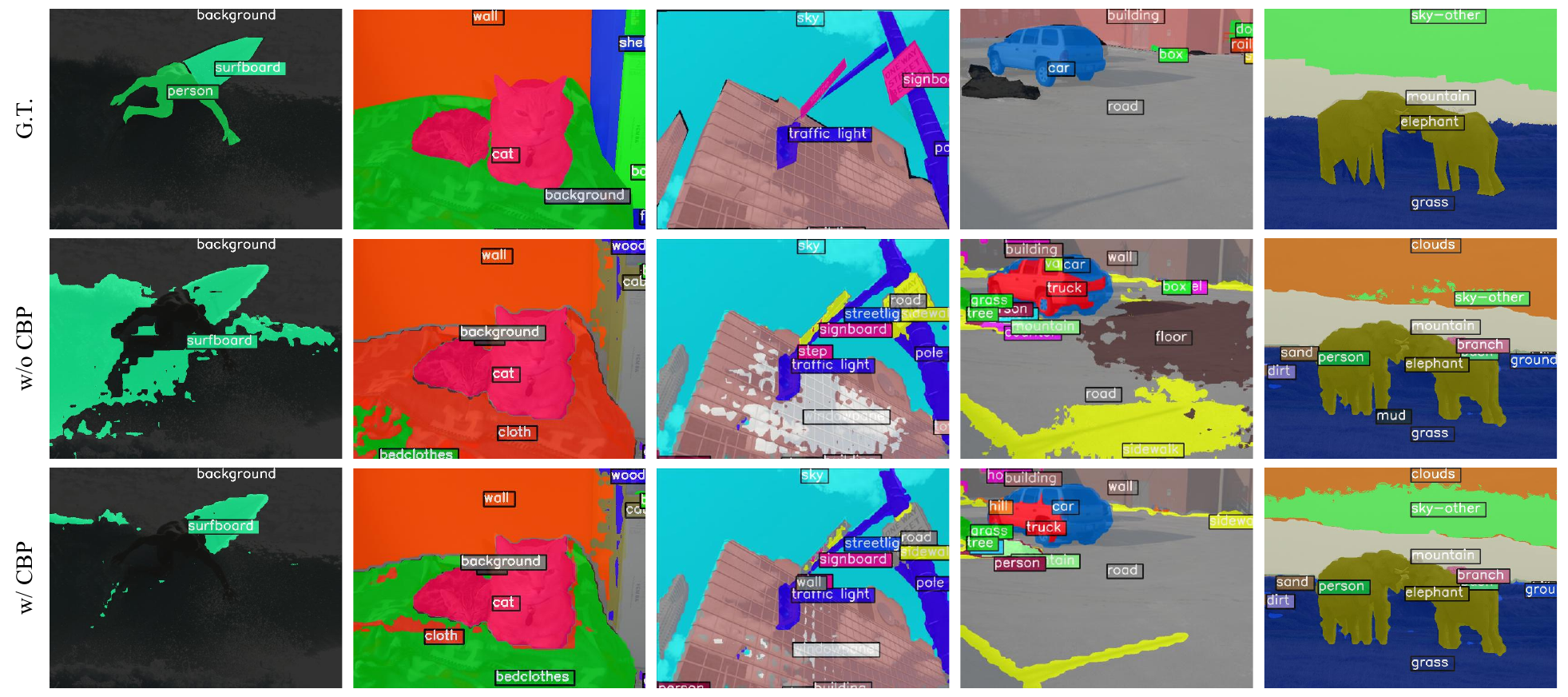}
    \caption{\textbf{Qualitative comparison showing the effect of CBP.} Qualitative comparison showing the effect of CBP. To enable a more explicit comparison, post-processing is removed; while higher feature coherence can cause larger regions to be assigned to the wrong class, CBP reduces class ambiguity and helps maintain correct labeling.}
    \label{fig:sup-cbp_anal}
\end{figure*}

\noindent This prompt design leads to more detailed and discriminative text representations than conventional template-based prompts and provides richer cues for vision–language alignment. A qualitative comparison between using CBP and not using CBP is presented in Fig.~\ref{fig:sup-cbp_anal}, and a pseudo-code illustrating how CBP is applied is shown in Algorithm~\ref{algo:cbp_alg}. Representative examples of the generated descriptions are provided in Tab.~\ref{tab:sup-CBP_expamples}.

\section{Additional Visualization Results.}
\begin{figure*}[t]
    \centering
    \begin{minipage}{0.9\textwidth}
        \includegraphics[width=\linewidth]{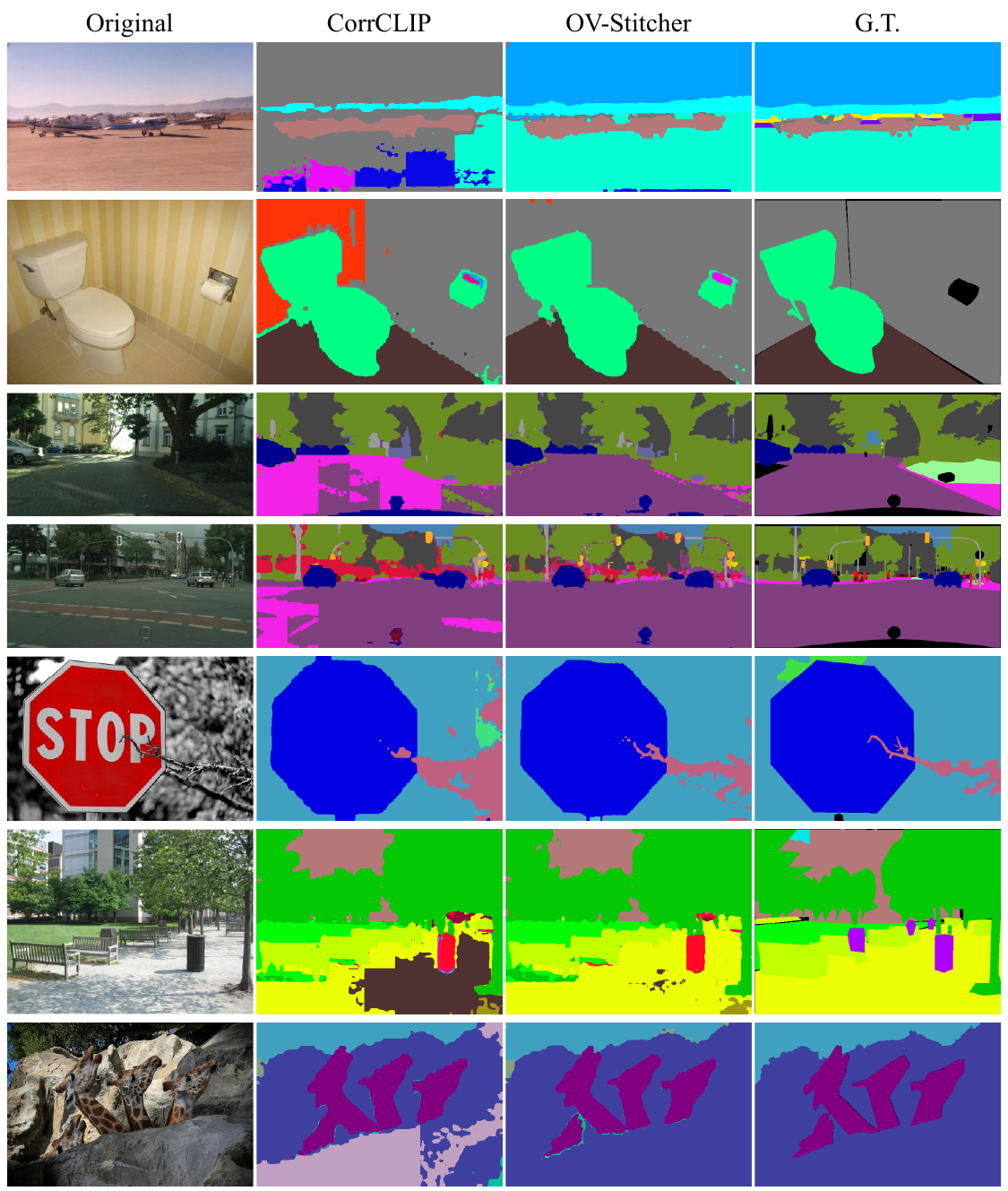}
        \caption{\textbf{Qualitative comparison without post-processing.} By removing post-processing, it becomes clear that our method produces more spatially and semantically feature-coherent results than the baseline CorrCLIP.}
        \label{fig:sup-no_mc}
    \end{minipage}
\end{figure*}

\begin{figure*}
    \centering
    \begin{minipage}{0.8\textwidth}
        \includegraphics[width=\linewidth]{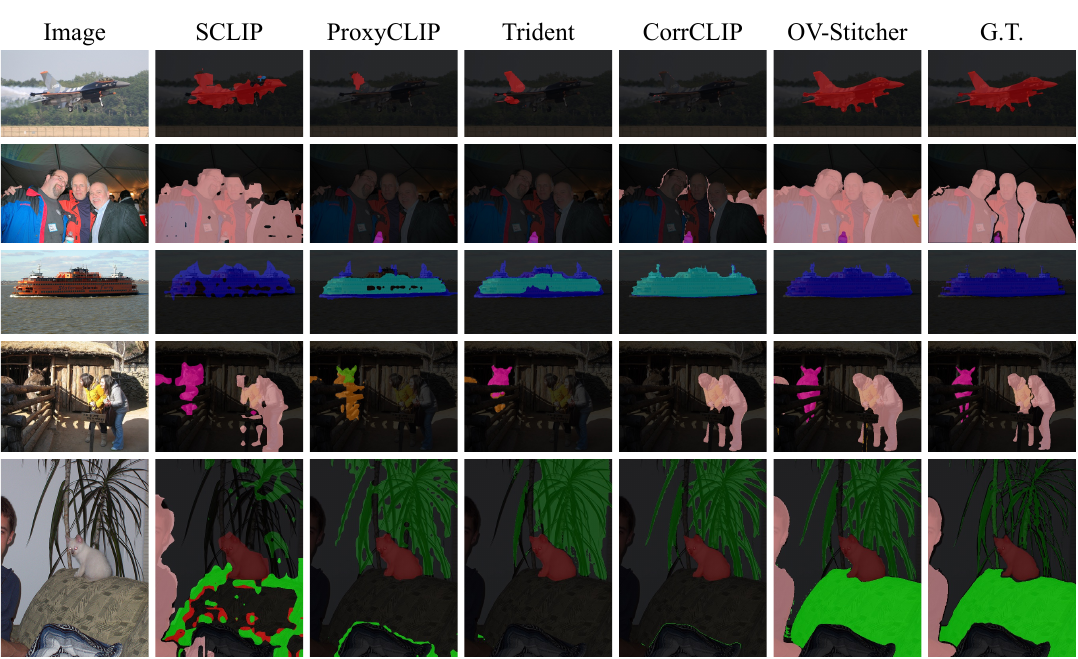}
        \caption{\textbf{Additional qualitative comparison on VOC21~\cite{pascal_voc}.}}
        \label{fig:v21_vis_supp}
    \end{minipage}
\vspace{-0.5cm}
\end{figure*}

We provide additional qualitative comparisons. Fig.~\ref{fig:sup-no_mc} shows a comparison between our method and the baseline CorrCLIP without post-processing, clearly illustrating the effectiveness of our approach. From Fig.~\ref{fig:v21_vis_supp} onward, we present the main qualitative results that include post-processing, comparing our method against previous approaches such as SCLIP~\cite{sclip}, ProxyCLIP~\cite{proxyclip}, Trident~\cite{trident}, and CorrCLIP~\cite{corrclip} across various datasets, including VOC21~\cite{pascal_voc}, Context60~\cite{pascal_context}, Cityscapes~\cite{cityscapes}, ADE20K~\cite{ade20k}, COCO Stuff~\cite{coco_stuff}, and COCO Object~\cite{coco_obj}.

\begin{algorithm}
   \caption{PyTorch-Like Code for Text Embeddings Generation}
   \label{algo:cbp_alg}
    \definecolor{codeblue}{rgb}{0.25,0.5,0.5}
    \lstset{
      basicstyle=\linespread{1.15}\fontsize{7.2pt}{7.2pt}\ttfamily,
      commentstyle=\fontsize{7.2pt}{7.2pt}\color{codeblue},
      keywordstyle=\color{red}\fontsize{7.2pt}{7.2pt},
    }
\begin{lstlisting}[language=python, numbers=none]
# cls_list: list of class names to embed
# CBP: dict that stores class-biased prompts for some classes
# CBP_generator: function that generates biased prompts when missing
def generate_text_embeddings(cls_list, CBP, CBP_generator)
  text_embeddings = []
  
  # (1) get class-biased prompts
  for cls in cls_list:
    if cls in CBP.keys():
      biased_prompt = CBP[cls]
    else:
      biased_prompt = CBP_generator(cls)
  
  # (2) build full prompt set: ImageNet templates + biased prompts
    prompts = [temp.format(cls) for temp in imagenet_temp] + biased_prompt
    query = tokenizer(prompts)
  
  # (3) encode prompts with CLIP text encoder
    feature = clip.encode_text(query)
    feature /= feature.norm(dim=-1, keepdim=True)
    feature = feature.mean(dim=0)
    feature /= feature.norm()
  
  # (4) store class embedding
    text_embeddings.append(feature.unsqueeze(0))
  
  # (5) stack all class embeddings
  text_embeddings = torch.cat(query_features, dim=0)

  return text_embeddings
\end{lstlisting}
\end{algorithm}

\begin{table*}[t]
\centering
\begin{tabular}{|c|p{12cm}|}
\hline
\textbf{Class} & \textbf{Class-Biased Prompts Examples} \\
\hline
\multirow{15}{*}{\centering goldfish} &
``a small, rounded body covered in shimmering scales''\newline
``a fish with a flat tail and vertical fins''\newline
``a small, orange fish with a white belly''\newline
``a gold-colored fish with a transparent tail''\newline
``a small, slender fish with a rounded head''\newline
``a fish with a long, flowing fins''\newline
``a small, yellow-gold fish with a black spot''\newline
``a fish with a flat, broad head''\newline
``a small, streamlined fish for swimming''\newline
``a fish with a bright orange dorsal fin''\newline
``a small, rounded fish with a horizontal stripe''\newline
``a fish with transparent scales reflecting light''\newline
``a small, gold-colored fish with a pointed snout''\newline
``a fish with a long, pointed fin on its back''\newline
``a small, slender fish with a distinctive pattern''
\\
\hline
\multirow{15}{*}{\centering salad} &
    ``a mix of leafy greens and colorful vegetables",\newline
    ``a bowl of fresh greens and vegetables arranged",\newline
    ``a tossed salad with mixed vegetables and greens",\newline
    ``a colorful salad with edible flowers",\newline
    ``a crunchy salad with crispy vegetables and nuts",\newline
    ``a fresh mix of lettuce and other leafy greens",\newline
    ``a salad with a variety of textures and colors",\newline
    ``a salad bowl filled with mixed greens and toppings",\newline
    ``a simple salad of mixed greens and cherry tomatoes",\newline
    ``a large salad bowl with multiple layers",\newline
    ``a colorful salad with fruits and vegetables",\newline
    ``a green salad with croutons and cheese",\newline
    ``a mixed salad with crunchy and soft ingredients",\newline
    ``a salad with a variety of leafy greens and vegetables",\newline
    ``a refreshing salad with lettuce, tomatoes, and cucumbers"
\\
\hline
\end{tabular}
\caption{\textbf{Representative examples of class-biased prompts generated for each category.}}
\label{tab:sup-CBP_expamples}
\end{table*}

\clearpage
\clearpage

\begin{figure*}
    \centering
    \begin{minipage}{0.8\textwidth}
        \includegraphics[width=\linewidth]{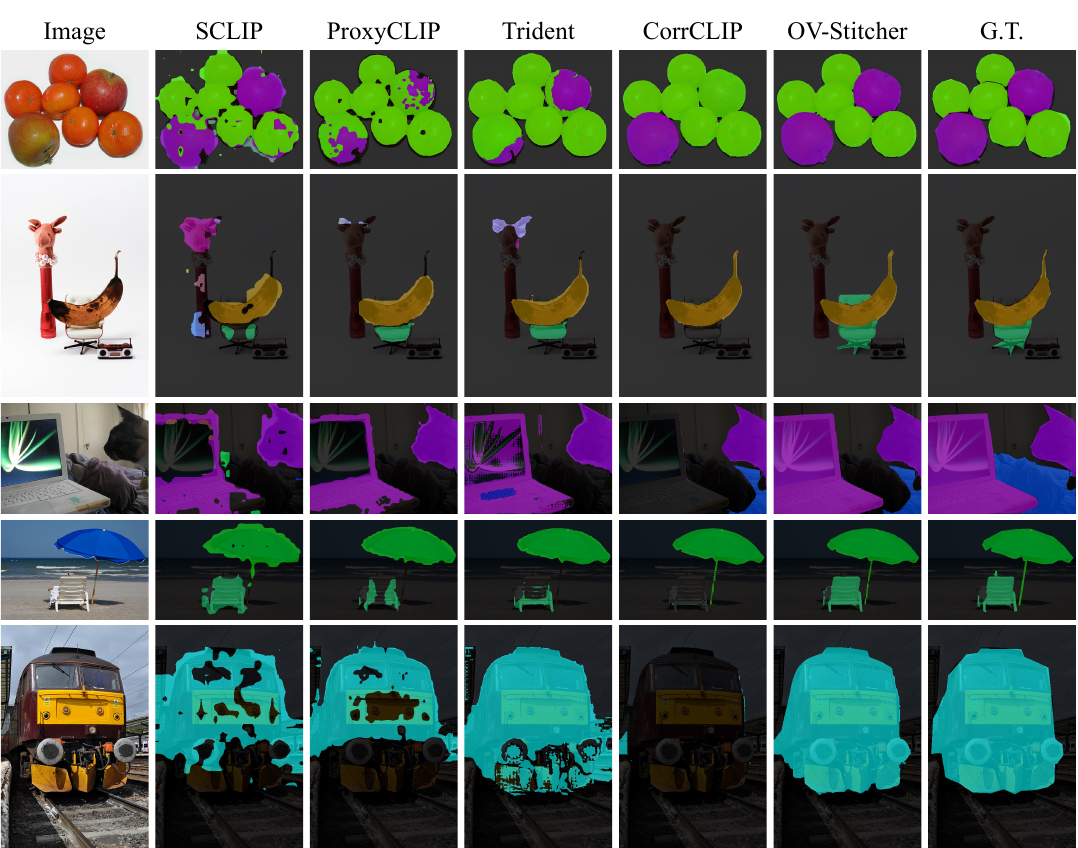}
        \caption{\textbf{Additional qualitative comparison on COCO Object~\cite{coco_stuff}.}}
        \label{fig:obj_vis_supp}
    \end{minipage}
\end{figure*}

\clearpage
\begin{figure*}
    \centering
    \begin{minipage}{0.9\textwidth}
        \includegraphics[width=\linewidth]{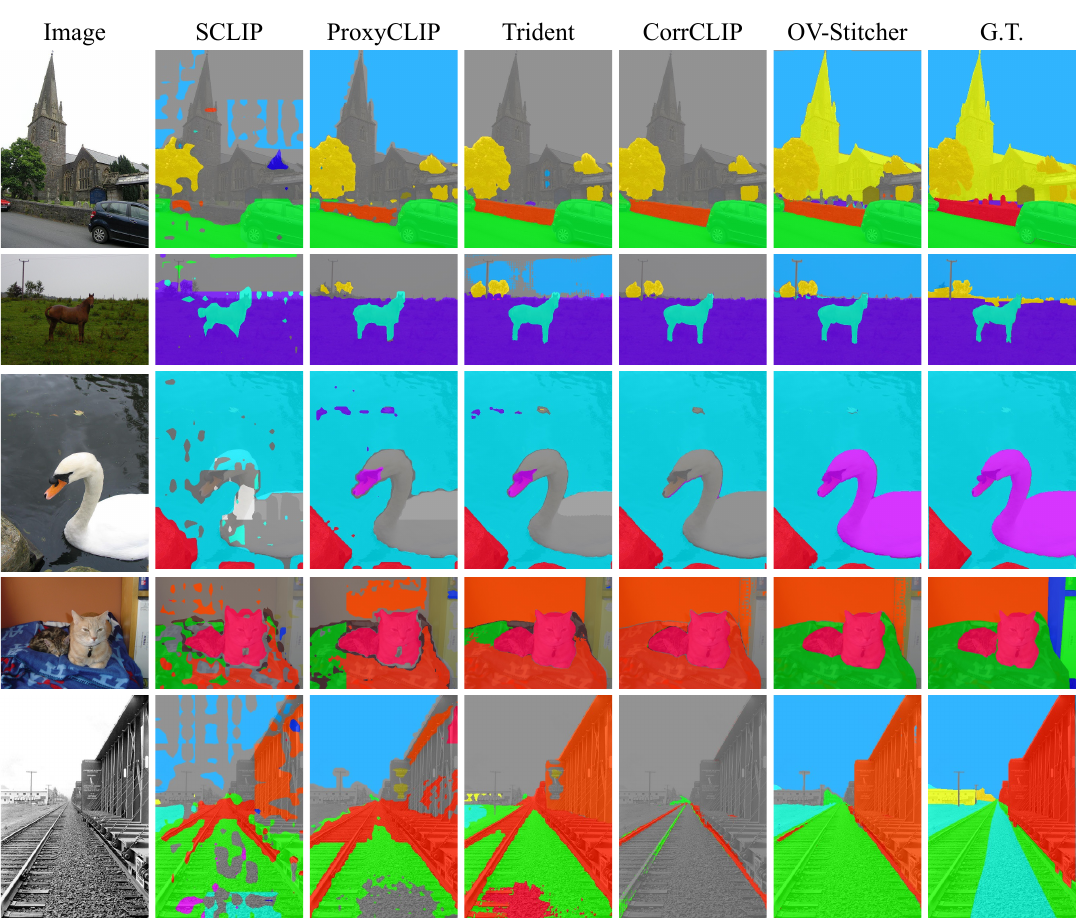}
        \caption{\textbf{Additional qualitative comparison on Context60~\cite{pascal_context}.}}
        \label{fig:c60_vis_supp}
    \end{minipage}
\end{figure*}

\begin{figure*}
    \centering
    \begin{minipage}{0.9\textwidth}
        \includegraphics[width=\linewidth]{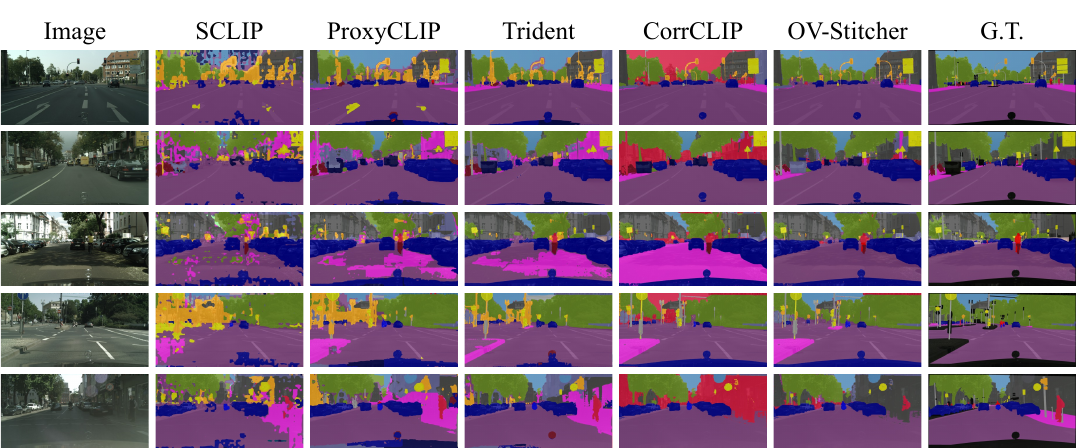}
        \caption{\textbf{Additional qualitative comparison on Cityscapes~\cite{cityscapes}}}
        \label{fig:cs_vis_supp}
    \end{minipage}
\end{figure*}
\clearpage
\begin{figure*}
    \centering
    \begin{minipage}{0.9\textwidth}
        \includegraphics[width=\linewidth]{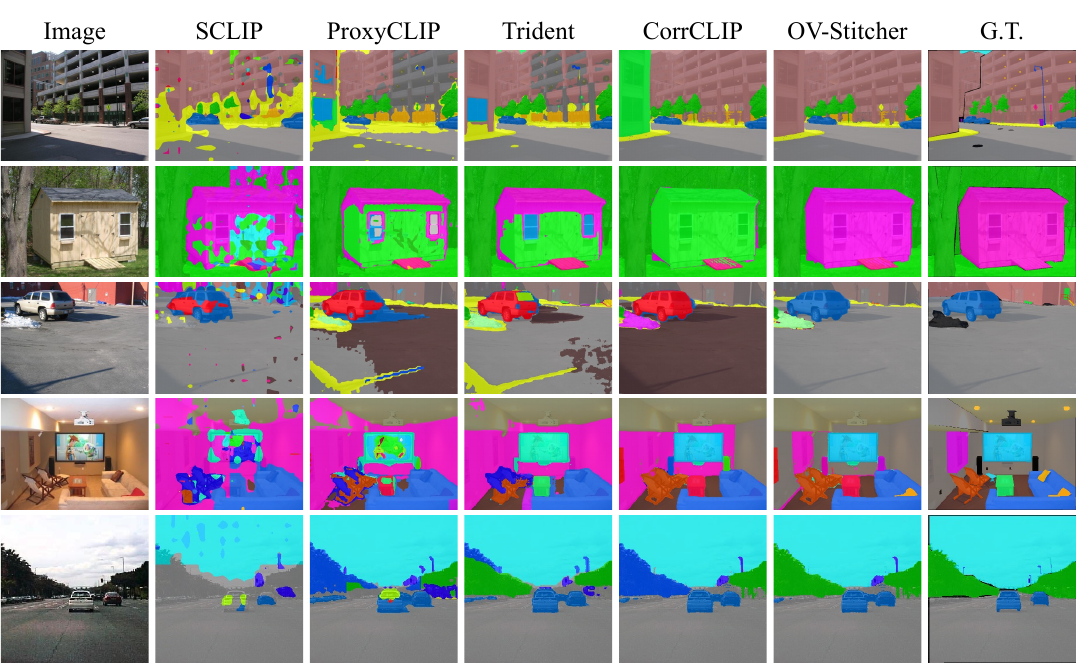}
        \caption{\textbf{Additional qualitative comparison on ADE20K~\cite{ade20k}}}
        \label{fig:ade_vis_supp}
    \end{minipage}
\end{figure*}

\begin{figure*}
    \centering
    \begin{minipage}{0.9\textwidth}
        \includegraphics[width=\linewidth]{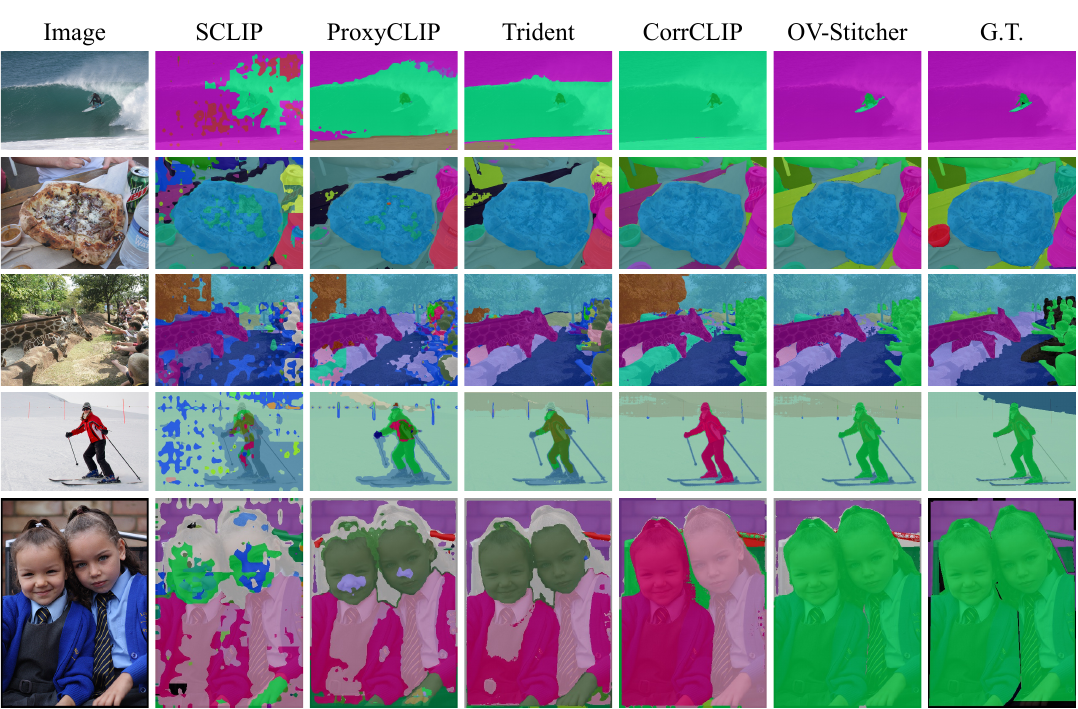}
        \caption{\textbf{Additional qualitative comparison on COCO Stuff~\cite{coco_stuff}}}
        \label{fig:stuff_vis_supp}
    \end{minipage}
\end{figure*}
\clearpage

\end{document}